\documentclass{article} 
\usepackage{iclr2027_conference,times}
\usepackage[T1]{fontenc}
\usepackage{newunicodechar}
\newunicodechar{“}{\textquotedblleft}
\newunicodechar{”}{\textquotedblright}

\usepackage{hyperref}
\usepackage{url}
\usepackage{amsmath,amssymb,bm,booktabs}
\usepackage{graphicx}
\usepackage{multirow}
\usepackage{wrapfig}
\usepackage{needspace}
\usepackage{array}
\usepackage{algorithm}
\usepackage{algpseudocode}
\usepackage{enumitem}
\usepackage{microtype}
\usepackage{longtable}
\usepackage{xcolor}
\usepackage{float}
\newcommand{\best}[1]{\textcolor[HTML]{D7191C}{\textbf{#1}}}
\newcommand{\sbest}[1]{\textcolor[HTML]{2C5FBF}{#1}}

\allowdisplaybreaks[4]

\renewcommand{\arraystretch}{0.95}

\title{SheafPrune: Sheaf-Guided Graph Propagation for Visual Token Pruning}

\author{
  Xiaoyang Ma\footnotemark[1] \\
  Key Lab of Smart Agriculture Systems, Ministry of Education\\
  China Agricultural University\\
  Beijing, 100083, China \\
  \texttt{s20243081931@cau.edu.cn}
  \And
  Chi Wang\thanks{These authors contributed equally to this work.} \\
  PHOENIX Lab\\
  Hangzhou Dianzi University\\
  Hangzhou, 310018, China \\
  \texttt{conanben77@hdu.edu.cn} \\
  \And
  Hong Sun\thanks{Corresponding author. This work was supported by Central Guidance Fund for Local Scientific and Technological Development Projects in Inner Mongolia (2024ZY0145)} \\
  Key Lab of Smart Agriculture Systems, Ministry of Education\\
  China Agricultural University\\
  Beijing, 100083, China \\
  \texttt{sunhong@cau.edu.cn}
}

\iclrfinalcopy 
\begin{document}

\maketitle

\begin{abstract}
Multimodal large language models process the complete visual token sequence, increasing prefill cost even when many tokens are irrelevant to the instruction. Token-wise saliency captures limited inter-token dependence; graph-based selection instead depends on how relations govern importance propagation and ranking. We study support-constrained propagation with node-degree correction. Inspired by sheaf geometry, SheafPrune uses modality-specific maps and a sparse text--visual graph to transfer relevance, then propagates scores on a separate visual graph and corrects them for degree. It retains a hard top-$k$ subset once before the frozen language model. The selector has about $4.5$M parameters and is trained end-to-end through the frozen LLaVA-NeXT-8B backbone using only the language-modeling objective. Explicit retention-ratio conditioning lets one checkpoint serve every budget. Across four backbones, ten benchmarks and seven pruning ratios, with all baselines re-implemented under one protocol, SheafPrune achieves the best retention at  $22$ of $28$ backbone--budget settings, with monotonically growing margins in the moderate-to-aggressive regime. On LLaVA-NeXT, retaining $10\%$ of visual tokens yields a $5.15\times$ net prefill speedup and an $84.0\%$ end-to-end FLOPs reduction. Together with component ablations, these results support graph-constrained importance propagation and degree correction as a design for visual token selection across the evaluated backbones and tokenization schemes, improving the trade-off between accuracy retention and inference cost.
\end{abstract}

\section{Introduction}
\label{sec:intro}

Multimodal large language models (MLLMs) encode an image into visual tokens and feed the visual sequence into a language model through a projector. Visual encoding covers the image independently of the current instruction, so redundant tokens pass through every decoder layer alongside tokens that carry essential evidence. Which tokens are redundant, however, depends on their relations and the question: a region dispensable for holistic recognition may contain an attribute or spatial cue; visually similar instances may jointly answer a counting question; a low-saliency region may carry the required text. A selector must therefore weigh tokens jointly under a finite retention budget, preserving useful evidence rather than simply keeping the largest individual scores.

Existing selectors draw on visual redundancy, language-model attention, or task-conditioned relevance. Graph methods, including G-Prune \citep{jiang2025gprune}, additionally relate visual tokens and propagate importance between them. For these methods, the initial scores do not determine the final ranking on their own. The graph specifies which tokens can exchange scores, while its connectivity affects how much each token receives. This motivates our central question: \emph{under a fixed visual-token budget, how does graph connectivity turn initial importance into an effective ranking, and what roles do propagation constraints and degree correction play?}
Answering this question requires resolving two key design issues. First, before relevance is defined, the states of visual tokens and the instruction, which come from different modalities, must be aligned so that they can be compared \citep{liang2022mind}. Second, propagation must be constrained to specified relations, and the effect of neighborhood size must be corrected; otherwise, densely connected tokens can attain higher ranks through connectivity alone \citep{kipf2017gcn,tang2020degree}.

Sheaf geometry addresses both issues through heterogeneous comparison and relation-constrained transfer: local states are compared after mapping them into shared edge spaces, and interactions follow specified relations \citep{hansen2019spectral,bodnar2022neural}. Following this perspective, \textbf{SheafPrune} forms an initial score from cross-modal relevance and a visual prior, propagates it over a thresholded visual graph, corrects it for neighborhood degree so that connectivity alone does not inflate a token's rank, and retains a hard Top-$k$ subset before the decoder. 
Modality-specific maps provide the comparison space; the visual stage uses an explicit graph transition rather than a learned sheaf Laplacian. Approximately $4.5$M selector parameters are trained through a frozen backbone. Retention-ratio conditioning enables \emph{one trained checkpoint to serve all evaluated budgets}, without training a separate selector for each budget.

Across four backbones and ten benchmarks, the reported per-point best-$\delta$ evaluation yields the highest mean retention at $22$ of $28$ backbone--budget settings. These are performance-envelope results; the fixed-configuration ablations answer a different question about the ranking operator. They show that uniform initialization preserves most of the full method's performance, whereas removing degree correction or changing the propagation operator substantially reduces retention. Selection also has a cost: on LLaVA-NeXT, retaining $10\%$ of candidates yields a $5.15\times$ net prefill speedup, but at less aggressive budgets some simpler selectors are faster. Our contributions are:
\begin{itemize}
\item A sheaf-guided visual-token selector combining cross-modal comparison, support-constrained propagation and degree-corrected ranking, with one trained checkpoint covering multiple retention budgets.
\item An analysis separating source-wise transition normalization from destination-wise degree correction, supported by fixed-configuration ablations across three budgets that identify the contribution of the combined ranking operator.
\item An evaluation of accuracy retention and inference cost across four backbones, together with explicit hyperparameter-selection scope and coverage diagnostics that characterize limits of score-based selection.
\end{itemize}

\begin{figure}[t]
\centering
\includegraphics[width=\textwidth]{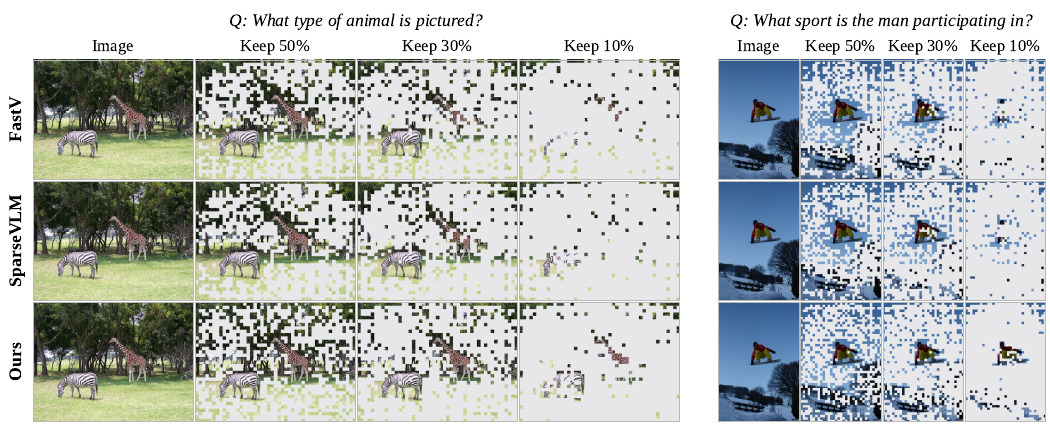}
\caption{\textbf{Which regions survive pruning.} Two VQAv2 examples on LLaVA-NeXT-8B at retention ratios $\rho\in\{0.5,0.3,0.1\}$. FastV, SparseVLM and SheafPrune retain the same number of candidates; pruned positions are greyed out. The examples illustrate differences in retained regions, rather than establish the causal effect of question conditioning. Further samples appear in Appendix~\ref{app:qualitative}.}
\label{fig:qualitative}
\end{figure}

\section{Related Work}
\subsection{Visual Token Pruning for MLLMs}
Visual compression methods differ in both their scoring signal and the operation applied to unselected tokens. VisionZip combines visual importance with feature similarity \citep{yang2025visionzip}, DivPrune emphasizes subset diversity \citep{alvar2025divprune}, and ToMe merges similar tokens \citep{bolya2023tome}. FastV uses shallow language-model attention \citep{chen2024fastv}; SparseVLM introduces textual signals \citep{zhang2025sparsevlm}; PyramidDrop removes tokens in stages \citep{xing2025pyramiddrop}. G-Prune propagates visual importance over a similarity graph \citep{jiang2025gprune}, making it a direct comparison for graph-based ranking. Instruction- and interaction-aware methods include IVTP, LVPruning and ATP-LLaVA \citep{huang2024ivtp,sun2025lvpruning,ye2025atpllava}, while CDPruner and CoIn study relevance and representativeness at the subset level \citep{zhang2025cdpruner,du2026coin}. Class-attention methods include LLaVA-PruMerge and FasterVLM \citep{shang2025prumerge,zhang2025fastervlm}. SheafPrune trains a small selector while keeping the backbone frozen. Our focus is how the graph operator transforms initial scores into a ranking; the experiments compare the evaluated implementations and do not establish superiority over untested selectors.

\subsection{Sheaf Theory in Machine Learning}
Cellular sheaves associate vector spaces with nodes and edges and use linear restriction maps to compare local states \citep{hansen2019spectral}. Neural Sheaf Diffusion learns these maps for graph representation learning \citep{bodnar2022neural}; knowledge sheaves use typed comparisons for relational data \citep{gebhart2023knowledge}. SheafPrune draws on this perspective to compare text and visual states in a shared space and to constrain information transfer to selected relations. Appendix~\ref{app:sheaf_background} gives a conditional cellular-sheaf interpretation of the cross-modal edge energy. The subsequent visual ranking uses an ordinary graph transition and count-based correction. The contribution concerns their construction and evaluation for token selection, rather than a new general sheaf-diffusion operator.

\section{Method}
\label{sec:method}
\subsection{Problem Formulation}
\label{sec:method_overview}
Let $X_T\in\mathbb R^{N_t\times d}$ and $X_V\in\mathbb R^{N_v\times d}$ denote text and visual representations at the selector interface. Masks $\bm u\in\{0,1\}^{N_t}$ and $\bm v,\bm c\in\{0,1\}^{N_v}$ identify valid text, valid visual positions, and prunable candidates, with $c_i\leq v_i$. Structural positions, such as image-newline tokens, have $v_i=1,c_i=0$ and are retained outside the candidate budget. For $\mathcal C=\{i:c_i=1\}$ and retention ratio $\rho$, the selector keeps exactly $k=\min\{|\mathcal C|,\max(1,\lceil\rho|\mathcal C|\rceil)\}$ candidates, plus mandatory structural positions. Appendix~\ref{app:notation_masks} defines the masks and empty-set handling. The notation $d$ denotes the selector input width, not an assertion that every backbone has the same native hidden dimension.

\subsection{Sheaf-Guided Design}
\label{sec:method_preliminaries}
Two ideas from sheaf geometry guide the construction. First, a node carries a local representation, while an incident edge provides a space in which neighboring states can be compared through restriction maps. For text and visual tokens, type-specific maps therefore express compatibility without requiring the original feature spaces to share the same coordinates. Second, graph incidences specify which local comparisons and transfers are admitted. SheafPrune realizes these ideas through modality-specific comparison and sparse text--vision relevance transfer, followed by propagation on a separate visual graph. For fixed input-dependent normalizers, the cross-modal maps define linear restrictions and their squared disagreement is a sheaf edge energy (Appendix~\ref{app:sheaf_background}). This correspondence motivates the comparison stage; the visual transition below transports scalar scores, and degree correction converts those scores into a ranking. Thus, \emph{sheaf-guided} identifies the design rationale and its concrete realization without identifying the two stages with one sheaf Laplacian.

\begin{figure}[!htbp]
\centering
\includegraphics[width=\textwidth]{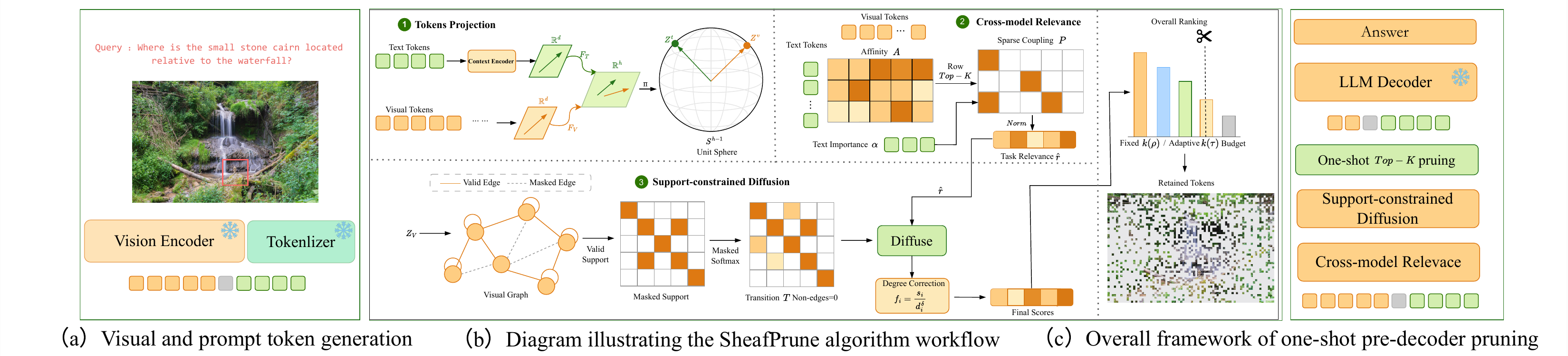}
\caption{Overview of \textbf{SheafPrune}. (a) Frozen encoders supply visual and textual representations. (b) Modality-specific projection and sparse cross-modal transfer produce relevance; support-constrained visual propagation and degree correction transform the fused initial score into a ranking. (c) Hard Top-$k$ selection physically removes unselected visual tokens before the frozen decoder. Matrices and scores are schematic.}
\label{fig:sheafprune_framework}
\end{figure}

\subsection{Budget-Conditioned Initialization}
\label{sec:method_text_context}
The textual mask admits the system prompt, completed dialogue turns and current question, but excludes target-answer positions, future turns, image placeholders, padding and control tokens. A lightweight bidirectional encoder produces $\widetilde X_T=\operatorname{Enc}_{\phi}(X_T;\bm u,\rho)$ outside the frozen language model, incorporating token order and the retention-ratio condition $g(\rho)$ (Appendix~\ref{app:text_encoder}). This keeps the selector's information boundary the same at training and inference.

After modality-specific LayerNorm, linear maps $F_T,F_V$ project tokens into an $h$-dimensional comparison space. Write $\widehat z=z/(\|z\|_2+\epsilon)$ for the stabilized normalized projection. Cross-modal affinity is $A_{ji}=\exp(-E_{ji}/\tau_E)$, with $E_{ji}=\|\widehat z_j^t-\widehat z_i^v\|_2^2\in[0,4]$ and learned $\tau_E>0$. For exact unit vectors, this energy is $2-2\langle\widehat z_j^t,\widehat z_i^v\rangle$; Appendix~\ref{app:cross_modal} distinguishes that identity from stabilized normalization. The modality-specific maps allow text and visual states to be transformed differently before comparison, while normalization controls their scale; semantic usefulness is learned through the downstream objective.

A masked softmax gives textual importance $\alpha$. Each valid text row retains its $K$ strongest valid visual neighbors; degree normalization yields $P=D_T^{-1/2}\widetilde A D_V^{-1/2}$ and relevance $\widehat r=\operatorname{Norm}_{\bm v}(P^\top\alpha)$. Invalid positions are excluded before neighbor selection. The initial score fuses relevance with the normalized visual-feature norm $\overline n$:
\begin{equation}
s^{(0)}=(1-\lambda)\overline n+\lambda\operatorname{std}_{\mathcal C}(\overline n)\operatorname{zscore}_{\mathcal C}(\widehat r),\qquad \lambda\in(0,1).
\label{eq:method_initialization}
\end{equation}
The scale factor matches the relevance term to the visual prior. The prior supplies a visual-side signal, while the centered relevance term can promote or demote candidates relative to it. These are initial importance scores rather than probabilities; graph propagation acts on their relative values, and final selection uses their ordering. Question information enters through $\widehat r$, with the downstream graph determining how that information influences neighboring tokens.

\subsection{Support-Constrained Propagation and Ranking}
\label{sec:method_cross_modal}
The visual graph is built from cosine similarities of valid visual features. Let $\xi_{q_v}$ be their sample-wise $(1-q_v)$ quantile over off-diagonal pairs. With explicit self-loops,
\begin{align}
\mathcal N(i)&=\{i\}\cup\{i'\ne i:v_{i'}=1,\operatorname{sim}_{ii'}\geq\xi_{q_v}\},\label{eq:method_visual_support}\\
T_{ii'}&=\frac{\mathbb1[i'\in\mathcal N(i)]\exp(\operatorname{sim}_{ii'})}{\sum_{l\in\mathcal N(i)}\exp(\operatorname{sim}_{il})}.\label{eq:method_transition}
\end{align}
Non-edges have exactly zero transition weight; retained edges remain valid even when their cosine similarity is negative. The global threshold gives symmetric support while allowing different neighborhood sizes. Scores are propagated as row vectors, $s^{(\ell+1)}=s^{(\ell)}T$, for $t$ steps, then corrected:
\begin{equation}
f_i=\frac{s_i^{(t)}}{d_i^\delta},\qquad d_i=|\mathcal N(i)|.
\label{eq:method_degree_correction}
\end{equation}

\textbf{Why normalization and correction differ.} Row normalization fixes the total weight leaving each source; it does not equalize the total weight received by each destination. With uniform initialization $s_j^{(0)}=c$, one step gives
\begin{equation}
s_i^{(1)}=c\sum_jT_{ji},\qquad f_i^{(1)}=\frac{c\sum_jT_{ji}}{d_i^\delta}.
\label{eq:method_uniform_step}
\end{equation}
Thus, graph connectivity can produce a nonuniform ranking even from uniform scores. Degree correction separately adjusts that ranking using neighborhood counts, not weighted row sums. This explains how the operations differ; their usefulness is tested in Section~\ref{sec:exp_ablation}. Appendix~\ref{app:visual_graph} analyzes longer-run degree bias and its limits for finite-step propagation. The two stages also use different measures of connectivity: $T$ depends on exponential similarity weights, whereas $d_i$ counts the retained neighbors. Dividing by $d_i^\delta$ therefore does not generally cancel the effects of propagation. Nodes with equal neighborhood counts can receive different propagated scores, while nodes with similar received mass can be reordered by correction. Neither operation explicitly enforces spatial coverage.

\begin{algorithm}[!htbp]
    \caption{SheafPrune selection. Stages are italicized; every statistic, softmax, and ranking is evaluated only on its valid support.}
    \label{alg:method_sheafprune}
    \small
    \begin{algorithmic}[1]
        \Require $X_T,X_V$; masks $\bm u,\bm v,\bm c$; retention ratio $\rho$
        \Ensure mask $m$; retained visual indices $\mathcal I_{\mathrm{keep}}$
        \State \textit{Budget:}\; $k\gets\min\{|\mathcal C|,\max(1,\lceil\rho|\mathcal C|\rceil)\}$;\;
               $\widetilde X_T\gets\operatorname{Enc}_\phi(X_T;\bm u,\rho)$
               \Comment{Eq.~\eqref{eq:app_budget},~\eqref{eq:app_text_output}}
        \State \textit{Relevance:}\; $Z_T,Z_V\gets$ modality maps;\;
               $A\gets\exp(-E/\tau_E)\odot\bm u\bm v^\top$
               \Comment{Eq.~\eqref{eq:app_affinity}}
        \State $\widetilde A\gets A\odot M_K$;\quad
               $\widehat r\gets\operatorname{Norm}_{\bm v}(P^\top\alpha)$
               \Comment{Eq.~\eqref{eq:app_bipartite_degrees},~\eqref{eq:app_bipartite_propagation}}
        \State \textit{Propagation:}\; $\mathcal N(i)\gets\{i\}\cup\{i'\!:v_{i'}=1,\operatorname{sim}_{ii'}\geq\xi_{q_v}\}$;\;
               $T\gets\operatorname{Softmax}_{\mathcal N(i)}(\operatorname{sim})$
        \State $s^{(0)}\gets(1-\lambda)\overline n+\lambda\operatorname{std}_{\mathcal C}(\overline n)\operatorname{zscore}_{\mathcal C}(\widehat r)$
               \Comment{Eq.~\eqref{eq:app_early_fusion}}
        \For{$\ell=0,\ldots,t-1$}\;
            $s^{(\ell+1)}\gets s^{(\ell)}T$
            \Comment{Eq.~\eqref{eq:app_diffusion_degree}}
        \EndFor
        \State $f_i\gets s_i^{(t)}/d_i^{\delta}$,\; $d_i=|\mathcal N(i)|$
               \Comment{degree correction}
        \State \textit{Selection:}\; $m\gets\operatorname{STTopK}(\operatorname{zscore}_{\mathcal C}(f),k;\beta)$
               \Comment{Eq.~\eqref{eq:method_training_gate}}
        \State $\mathcal I_{\mathrm{keep}}\gets$ selected candidates and mandatory positions in original order
        \State \Return $m,\mathcal I_{\mathrm{keep}}$;\; train: $p'\gets\operatorname{Gate}(p;m)$,\; infer: $X_V^{\mathrm{keep}}\gets\operatorname{Gather}(X_V,\mathcal I_{\mathrm{keep}})$
    \end{algorithmic}
\end{algorithm}

\subsection{Training and Inference}
\label{sec:method_training}
Candidate scores are standardized and selected with a straight-through Top-$k$ estimator \citep{bengio2013estimating}. The forward pass uses a hard subset, while the backward pass uses a smooth surrogate; hard selection itself is not differentiable. During training, the mask gates attention probabilities:
\begin{equation}
m=\operatorname{STTopK}(\operatorname{zscore}_{\mathcal C}(f),k;\beta),\qquad
p'_{q\kappa}=\frac{\mu_\kappa p_{q\kappa}}{\sum_{\kappa'}\mu_{\kappa'}p_{q\kappa'}+\epsilon}.
\label{eq:method_training_gate}
\end{equation}
Here $\mu_\kappa$ is the visual gate, or one for textual and mandatory structural positions. The selector is trained on LLaVA-NeXT-8B: the vision encoder, projector and language model remain frozen, and only the approximately $4.5$M selector parameters are updated by target-answer language-modeling loss. Gradients pass through the frozen generation path and the multiplicative gate with renormalization. Training uses the LLaVA-NeXT instruction mixture\citep{liu2024llavanext}, weighted by the signal its sources provide for visual-token selection. After initial training at $\rho=0.9$, retention-ratio sampling and conditioning through $g(\rho)$ train one checkpoint for all evaluated budgets (Appendix~\ref{app:training_setup}). The selector sees the same admissible context during training and inference; target-answer positions are excluded before its text encoder. Appendix~\ref{app:training} and Appendix~\ref{app:training_setup} specify the estimator, objective and training configuration.

At inference, the selector runs once before the decoder. Unselected visual tokens are physically removed, and retained tokens preserve their original position identifiers, including multidimensional coordinates where required by the backbone. Gating and physical removal share the same hard subset but remain distinct executions, so strict numerical equivalence is not assumed. All reported accuracy and efficiency results use the physical-removal path (Appendix~\ref{app:physical_pruning}).

\subsection{Space and Speed}
\label{sec:method_efficiency}
One-shot removal shortens the sequence from the first decoder layer. Its net benefit depends on whether saved decoder computation exceeds selector overhead. Materializing the text attention, cross-modal affinity and visual similarity matrices requires auxiliary space $\mathcal O(N_t^2+N_tN_v+N_v^2)$; sparse propagation alone does not eliminate the quadratic similarity workspace. All reported prefill latencies include selection cost. Training is an offline cost, while visual encoding and projection occur before pruning and are not avoided. For competing methods that also shorten the sequence once before the decoder, the remaining difference depends on selector kernels, memory access and execution paths. The accuracy--latency trade-off is therefore an empirical system result under the stated measurement protocol. Appendix~\ref{app:efficiency} gives the cost comparison.

\section{Experiments}
\label{sec:exp}
\subsection{Setup}
\label{sec:exp_setup}
\textbf{Backbones, benchmarks and metrics.} We evaluate LLaVA-NeXT-8B \citep{liu2024llavanext}, Qwen2.5-VL-7B \citep{bai2025qwen25vl}, InternVL3-8B \citep{zhu2025internvl3}, and LLaVA-1.5-7B \citep{liu2024llava15}. The ten benchmarks are ChartQA \citep{masry2022chartqa}, DocVQA \citep{mathew2021docvqa}, TextVQA \citep{singh2019textvqa}, HR-Bench 4K/8K \citep{wang2025hrbench}, GQA \citep{hudson2019gqa}, VQAv2 \citep{goyal2017vqav2}, MMBench-EN \citep{liu2024mmbench}, POPE \citep{li2023pope} and MME \citep{fu2025mme}.
These benchmarks cover text-rich chart and document understanding (ChartQA, DocVQA, TextVQA), fine-grained perception in high-resolution images (HR-Bench 4K/8K), compositional reasoning and general visual question answering (GQA, VQAv2), holistic multimodal perception and cognition (MMBench-EN, MME), and object hallucination (POPE).
All methods use the same benchmark splits and official metric definitions (Appendix~\ref{app:eval_protocol}). RA\% is the macro-average of ten pruned-to-unpruned score ratios; MME is normalized before aggregation. With pruning rate $p=1-\rho$, we evaluate $p\in\{0.1,0.3,0.5,0.6,0.7,0.8,0.9\}$. Prefill latency includes selection; LR\% compares it with the unpruned reference.

\textbf{Baselines.} We use the baselines' open-source implementations and published method specifications within a common evaluation pipeline. Random, G-Prune, FastV, SparseVLM, PyramidDrop and ToMe are evaluated on all four backbones; VisionZip, PruMerge and FasterVLM are reported on LLaVA-1.5. Appendix~\ref{app:baselines} describes evaluation coverage and layer-averaged budget accounting. A shared pipeline standardizes data and metrics, while algorithm-specific pruning locations and execution costs remain different. Claims are restricted to the evaluated methods and settings.

\begin{figure}[!htbp]
\centering
\includegraphics[width=\textwidth]{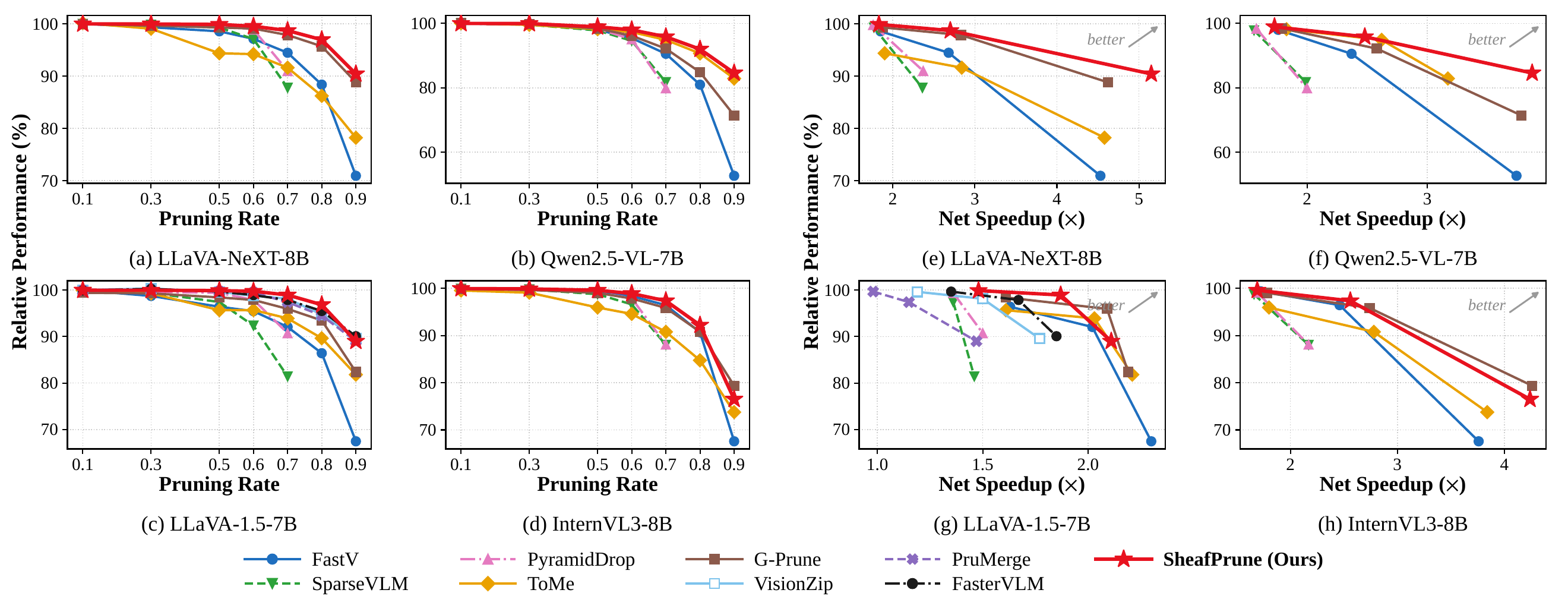}
\caption{\textbf{Reported best-$\delta$ performance envelope and prefill trade-offs.} RA\% averages ten benchmark retention ratios. The SheafPrune entries use per-point degree-exponent selection; they do not represent one fixed-$\delta$ deployment configuration. Unavailable staged-method points reflect the layer-averaged budget limits described in Appendix~\ref{app:baselines}.}
\label{fig:relative_performance}
\end{figure}

\textbf{Training and reporting protocol.} Training has two stages: approximately $0.3$ of an epoch on a strong-signal subset at $\rho=0.9$, followed by approximately $2$ epochs on the source-weighted $779$K mixture. Stage B repeats four steps sampling $\rho$ uniformly from $\{0.5,0.3,0.1\}$ followed by one additional step at $\rho=0.9$. The ratio is supplied through $g(\rho)$; one checkpoint serves all seven evaluation budgets and is reused for the other backbones without retraining. The main comparison reports the best result over $\delta\in\{0,0.5,1,2\}$ at each evaluation point: a performance envelope, not a fixed-$\delta$ deployment curve. The $22/28$ count and margin trends follow this protocol. Component ablations use a fixed default configuration, changing only the specified arm. Checkpoint reuse refers to trained weights; inference hyperparameters follow the stated reporting protocol. Appendix~\ref{app:training_setup} gives the sampling, optimization and stage configurations.

\subsection{Accuracy Retention and Inference Cost}
\label{sec:exp_main}
Figure~\ref{fig:relative_performance} summarizes the seven pruning rates; Table~\ref{tab:exp_main} reports per-benchmark scores at $p=0.7$, and Appendix~\ref{app:rawscores} gives all budgets. Under the reported best-$\delta$ protocol, SheafPrune leads all $16$ backbone--budget points at $p=0.5$--$0.8$. The reported margins over the strongest competitor increase across this range, from $+0.10$ to $+1.25$ on LLaVA-NeXT and $+0.22$ to $+1.40$ on InternVL3. These are comparisons of point estimates, not significance claims. Across all four backbones, the advantage is narrow at light pruning and widens consistently as the budget tightens.

\begin{table}[!htbp]
    \caption{Comparison across four backbones (SheafPrune: reported best-$\delta$ protocol) at $p=0.7$ ($\rho=0.3$).
    acc: benchmark score (MME/28); lat: prefill latency including selection (ms); RA\%: ten-benchmark macro-average retention; LR\%: latency reduction, both against the unpruned backbone.
    Red/blue mark the best/second-best pruning method per column and backbone; Appendix~\ref{app:eval_protocol} gives the profiling protocol and Table~\ref{tab:exp_all} the native scores and the remaining budgets.}
    \label{tab:exp_main}
    \centering
    \setlength{\tabcolsep}{1.5pt}
    \renewcommand{\arraystretch}{0.95}
    \scriptsize
    \begin{tabular}{@{}ll cc|ccc|ccc|cc|cc|cc@{}}
        \toprule
        & & \multicolumn{2}{c|}{General VQA} & \multicolumn{3}{c|}{MLLM benchmarks} & \multicolumn{3}{c|}{Text-oriented VQA} & \multicolumn{2}{c|}{High-resolution} & \multicolumn{2}{c|}{Avg} & & \\
        \cmidrule(lr){3-4}\cmidrule(lr){5-7}\cmidrule(lr){8-10}\cmidrule(lr){11-12}\cmidrule(lr){13-14}
        Model & Method & GQA & VQAv2 & MME & POPE & MMB & ChartQA & DocVQA & TextVQA & HR-4K & HR-8K & acc & lat & RA\% & LR\%\\
        \midrule
        \multirow{7}{*}{\rotatebox[origin=c]{90}{LLaVA-NeXT-8B}} & No prune & 65.0 & 82.1 & 38.4 & 83.6 & 77.0 & 69.7 & 72.6 & 66.7 & 51.2 & 46.1 & 65.2 & 149.4 & -- & --\\
        \cmidrule(l){2-16}
         & FastV & 63.5 & 81.0 & 37.7 & 82.5 & 76.1 & 59.2 & 55.2 & 61.7 & \sbest{51.4} & 45.5 & 61.4 & 55.7 & 94.5 & 62.7\\
         & SparseVLM & 58.5 & 76.5 & 37.1 & 83.0 & \best{76.8} & 45.0 & 46.4 & 46.9 & 51.0 & \best{46.1} & 56.7 & 63.3 & 87.8 & 57.6\\
         & PyramidDrop & 62.2 & 80.8 & 37.8 & \best{85.0} & 76.1 & 45.4 & 49.3 & 55.5 & \best{52.1} & 45.6 & 59.0 & 63.0 & 91.0 & 57.8\\
         & ToMe & 64.5 & 80.8 & 37.8 & 82.4 & \sbest{76.3} & 52.1 & 48.7 & 60.5 & 50.0 & 42.5 & 59.6 & \best{52.6} & 91.6 & \best{64.8}\\
         & G-Prune & \best{64.9} & \sbest{81.6} & \sbest{37.8} & \sbest{83.6} & 75.7 & \sbest{66.5} & \sbest{64.8} & \sbest{65.6} & 51.2 & 45.8 & \sbest{63.8} & \sbest{52.8} & \sbest{97.9} & \sbest{64.7}\\
         & \textbf{SheafPrune (Ours)} & \sbest{64.8} & \best{82.1} & \best{38.4} & 83.6 & 76.2 & \best{67.5} & \best{67.2} & \best{66.1} & 51.2 & \sbest{46.1} & \best{64.3} & 55.3 & \best{98.7} & 63.0\\
        \midrule
        \multirow{7}{*}{\rotatebox[origin=c]{90}{Qwen2.5-VL-7B}} & No prune & 58.7 & 83.1 & 44.4 & 84.3 & 83.3 & 87.8 & 94.2 & 84.3 & 71.4 & 63.9 & 75.5 & 113.6 & -- & --\\
        \cmidrule(l){2-16}
         & FastV & 51.7 & 76.8 & \sbest{41.8} & 79.9 & 79.5 & 63.0 & 74.5 & \best{82.3} & 68.1 & 61.9 & 68.0 & 47.9 & 90.6 & 57.8\\
         & SparseVLM & 49.2 & 68.9 & 40.2 & 76.2 & 76.9 & 27.3 & 67.3 & 66.7 & 69.5 & 63.0 & 60.5 & 57.1 & 81.8 & 49.7\\
         & PyramidDrop & 48.1 & 71.5 & 36.4 & 74.9 & 71.3 & 41.0 & 67.6 & 69.1 & 63.2 & 54.6 & 59.8 & 56.8 & 79.9 & 50.0\\
         & ToMe & \best{58.2} & \best{80.8} & \best{43.4} & \sbest{83.3} & \best{81.3} & \sbest{73.1} & 84.7 & \sbest{79.5} & 68.1 & 60.9 & \sbest{71.3} & \best{43.4} & \sbest{94.9} & \best{61.8}\\
         & G-Prune & 51.2 & 74.8 & 39.8 & 80.1 & 78.5 & 66.8 & \sbest{88.0} & 78.4 & \best{72.8} & \best{64.5} & 69.5 & \sbest{44.0} & 92.2 & \sbest{61.2}\\
         & \textbf{SheafPrune (Ours)} & \sbest{57.0} & \sbest{80.0} & 41.3 & \best{84.2} & \sbest{81.2} & \best{77.1} & \best{88.1} & 79.1 & \sbest{71.4} & \sbest{63.5} & \best{72.3} & 45.8 & \best{95.8} & 59.7\\
        \midrule
        \multirow{10}{*}{\rotatebox[origin=c]{90}{LLaVA-1.5-7B}} & No prune & 58.7 & 75.8 & 36.8 & 81.1 & 67.9 & 18.0 & 24.6 & 47.4 & 33.9 & 36.2 & 48.1 & 39.7 & -- & --\\
        \cmidrule(l){2-16}
         & FastV & 53.9 & 71.8 & 35.1 & 73.8 & 66.5 & 15.8 & 18.3 & 42.2 & 33.6 & 36.1 & 44.7 & 19.7 & 92.1 & 50.5\\
         & SparseVLM & 49.8 & 64.6 & 34.7 & 72.9 & 65.6 & 13.7 & 10.6 & 21.8 & \sbest{35.2} & 34.1 & 40.3 & 27.2 & 81.4 & 31.5\\
         & PyramidDrop & 55.3 & 72.1 & 34.9 & 73.5 & 65.6 & 16.0 & 17.1 & 40.4 & 33.1 & 34.4 & 44.2 & 26.5 & 90.7 & 33.3\\
         & ToMe & \sbest{57.4} & 73.9 & 36.0 & \best{79.7} & 66.2 & 16.0 & 17.8 & 42.7 & 34.4 & 35.6 & 46.0 & \sbest{19.6} & 93.9 & \sbest{50.7}\\
         & G-Prune & 55.1 & 72.9 & 36.2 & 75.5 & 65.2 & 17.1 & 22.6 & \sbest{46.5} & 33.1 & 36.1 & 46.0 & \best{19.0} & 96.0 & \best{52.2}\\
         & FasterVLM & 55.9 & 74.3 & 36.6 & 76.4 & 66.8 & 17.6 & 22.7 & 45.5 & \best{35.8} & \best{37.1} & 46.9 & 23.8 & 97.9 & 40.1\\
         & VisionZip & 56.5 & \sbest{74.4} & \best{36.6} & 76.9 & \sbest{67.0} & \sbest{17.7} & \sbest{23.1} & 45.6 & 35.1 & \sbest{37.1} & \sbest{47.0} & 26.5 & \sbest{98.1} & 33.3\\
         & LLaVA-PruMerge & 56.7 & 74.4 & \sbest{36.6} & 77.7 & 66.7 & 17.0 & 22.5 & 45.8 & 34.8 & 36.8 & 46.9 & 34.5 & 97.4 & 13.0\\
         & \textbf{SheafPrune (Ours)} & \best{57.7} & \best{74.9} & 36.4 & \sbest{79.3} & \best{67.3} & \best{17.9} & \best{24.2} & \best{47.1} & 33.9 & 36.2 & \best{47.5} & 21.2 & \best{99.0} & 46.5\\
        \midrule
        \multirow{7}{*}{\rotatebox[origin=c]{90}{InternVL3-8B}} & No prune & 61.3 & 79.3 & 44.8 & 89.6 & 87.0 & 86.5 & 92.5 & 82.3 & 70.0 & 61.9 & 75.5 & 128.7 & -- & --\\
        \cmidrule(l){2-16}
         & FastV & \sbest{60.5} & \best{79.1} & \best{44.3} & 88.7 & \best{86.2} & \sbest{76.5} & 75.5 & \best{81.4} & \sbest{70.4} & 61.5 & \sbest{72.4} & 52.3 & \sbest{96.4} & 59.3\\
         & SparseVLM & 58.3 & 76.4 & 43.7 & 87.5 & 84.9 & 45.2 & 60.8 & 66.5 & 68.8 & 61.2 & 65.3 & 59.3 & 88.0 & 53.9\\
         & PyramidDrop & 58.7 & 76.2 & 41.4 & 86.8 & 81.3 & 60.6 & 62.4 & 68.8 & 64.5 & 57.5 & 65.8 & 59.3 & 88.1 & 53.9\\
         & ToMe & 60.5 & 77.8 & 42.8 & 88.6 & 85.8 & 63.6 & 62.1 & 72.4 & 67.4 & 57.4 & 67.8 & \best{46.3} & 90.8 & \best{64.0}\\
         & G-Prune & 59.7 & 78.0 & \sbest{44.1} & \sbest{89.3} & 85.5 & 72.2 & \sbest{77.8} & 78.5 & \best{71.0} & \best{63.2} & 71.9 & \sbest{47.0} & 95.8 & \sbest{63.5}\\
         & \textbf{SheafPrune (Ours)} & \best{61.0} & \sbest{78.0} & 43.3 & \best{89.5} & \sbest{86.0} & \best{81.6} & \best{83.3} & \sbest{79.1} & 70.0 & \sbest{61.6} & \best{73.4} & 50.3 & \best{97.3} & 60.9\\
        \bottomrule
    \end{tabular}

\end{table}

\textbf{Performance and exceptions.} At $p=0.7$ on LLaVA-NeXT, DocVQA scores are $67.23$ for SheafPrune and $64.84$ for the strongest evaluated baseline. At $p=0.9$, SheafPrune has the highest reported RA on LLaVA-NeXT ($90.42\%$) and Qwen2.5-VL ($84.60\%$), but is overtaken by FasterVLM and VisionZip on LLaVA-1.5 and by G-Prune on InternVL3. It leads $22$ of $28$ settings overall under the envelope protocol. The four backbones span different tokenization schemes; the results support application across these evaluated interfaces.

\textbf{Accuracy has a selection cost.} At $p=0.7$ on LLaVA-NeXT, SheafPrune retains $98.71\%$ of performance, against $91.61\%$ for ToMe and $97.86\%$ for G-Prune, with net prefill speedups of $2.70\times$, $2.84\times$ and $2.83\times$, respectively. Better retention therefore costs about $5\%$ more latency than these two methods at this budget. At $p=0.9$, the reported accuracy and speed optima coincide on LLaVA-NeXT ($90.42\%$ RA, $5.15\times$ speedup) and Qwen2.5-VL ($84.60\%$, $3.87\times$), but not on the other two backbones. The reported selector overhead is $2.6\%$ of unpruned prefill on LLaVA-NeXT and $7.2\%$ on the shorter LLaVA-1.5 sequence, where the $p=0.9$ speedup is $2.11\times$. Across token counts and budgets, the selector costs at most a few percent of prefill while recovering several points of retention, and on the longer sequences it delivers the best accuracy and the best speedup at once.

\textbf{Pruning location and inference cost.} A fixed retained set shortens every decoder layer, whereas decoder-internal methods retain longer sequences in early layers. Under the budget accounting in Appendix~\ref{app:baselines}, nominal $90\%$ pruning for FastV corresponds to $16.4\%$ layer-averaged retention for $L=28,K=2$; the implemented staged schedules have a $25\%$ floor. This distinction motivates reporting the layer-averaged budget alongside accuracy. Prefill speed and end-to-end FLOPs remain different quantities: the $84.0\%$ FLOPs reduction on LLaVA-NeXT includes computation before pruning that cannot be removed and does not measure full-answer latency. At $p=0.7$, reported decoding throughput changes from $66.5$ to $68.3$ tokens/s, while the three decoder-internal methods report $50.9$--$51.7$ tokens/s. These are observations of the documented execution paths; Appendix~\ref{app:eval_protocol} gives the timing workload and profiling corrections.

\subsection{Ablation and Diagnosis}
\label{sec:exp_ablation}
\textbf{How the operator forms the ranking.} Table~\ref{tab:exp_ablation} compares six configurations on LLaVA-NeXT using the same defaults. At $p=0.9$, removing degree correction reduces RA from $89.39\%$ to $72.80\%$; Dense normalization gives $72.68\%$, against $73.38\%$ for random selection. In the documented dense variant, non-edge logits are zero before softmax and therefore contribute $\exp(0)=1$, changing both effective propagation support and the denominator (Appendix~\ref{app:component_ablation}). It is an internal operator ablation, not a description of the G-Prune baseline. The similar losses do not establish identical failure mechanisms. Degree-only ranking reaches $81.38\%$: useful degree information alone does not recover the full ranking. On the text-oriented tasks, Dense normalization and no degree correction lose $36.87$ and $36.67$ retention points relative to the full method; degree-only ranking remains $17.88$ points lower. The task-group columns therefore show a larger loss than the ten-task average, while the three budget columns show the pattern beyond a single pruning rate.

\begin{table}[!htbp]
\centering
\small
\setlength{\tabcolsep}{3.5pt}
\caption{\textbf{Fixed-configuration component ablation on LLaVA-NeXT-8B.} Defaults: $t=3$, $q_v=0.05$, $\delta=1$, except as indicated. Norm.: the pair set used for normalization; Text: mean retention on ChartQA, DocVQA and TextVQA. Subscripts are differences from the full method at $p=0.9$. RA averages ten benchmark retention ratios. These entries use fixed configurations, not the best-$\delta$ envelope; complete arms appear in Table~\ref{tab:app_ablation_full}.}
\label{tab:exp_ablation}
\begin{tabular}{@{}lccrrr@{}lr@{}l@{}}
\toprule
& & & \multicolumn{4}{c}{RA (\%)} & \multicolumn{2}{c}{Text}\\
\cmidrule(lr){4-7}\cmidrule(l){8-9}
Arm & Norm. & $\delta$ & $p=0.5$ & $p=0.7$ & \multicolumn{2}{c}{$p=0.9$} & \multicolumn{2}{c}{$p=0.9$}\\
\midrule
Random selection & --- & --- & 92.12 & 86.32 & 73.38 & $_{\,-16.01}$ & 39.20 & $_{\,-35.77}$\\
Degree only & --- & $1$ & 95.77 & 92.14 & 81.38 & $_{\,-8.01}$ & 57.09 & $_{\,-17.88}$\\
\midrule
Dense normalization & \textbf{dense} & $1$ & 91.74 & 85.78 & 72.68 & $_{\,-16.71}$ & 38.10 & $_{\,-36.87}$\\
No degree correction & support & $\mathbf{0}$ & 91.88 & 85.97 & 72.80 & $_{\,-16.59}$ & 38.30 & $_{\,-36.67}$\\
\midrule
Uniform initial score & support & $1$ & 98.50 & 96.72 & 88.19 & $_{\,-1.20}$ & 73.32 & $_{\,-1.65}$\\
\textbf{SheafPrune} & support & $1$ & \textbf{99.43} & \textbf{97.97} & \textbf{89.39} & & \textbf{74.97} &\\
\bottomrule
\end{tabular}
\end{table}

\textbf{What initialization adds.} Uniform initialization is  $0.93$, $1.25$ and $1.20$ RA points below the full method at the three budgets. This is consistent with Eq.~\eqref{eq:method_uniform_step}: connectivity can create a ranking without an initially nonuniform signal. It does not show that training is unnecessary or that questions never affect selection. Rather, the tested operator retains most of the observed benefit, and these ablations do not establish learned initialization as the main gain source. Removing only relevance or only the visual prior performs worse than uniform initialization; Appendix~\ref{app:component_ablation} reports those interactions without assigning statistical significance to small differences.

\begin{figure}[!htbp]
\centering
\includegraphics[width=\textwidth]{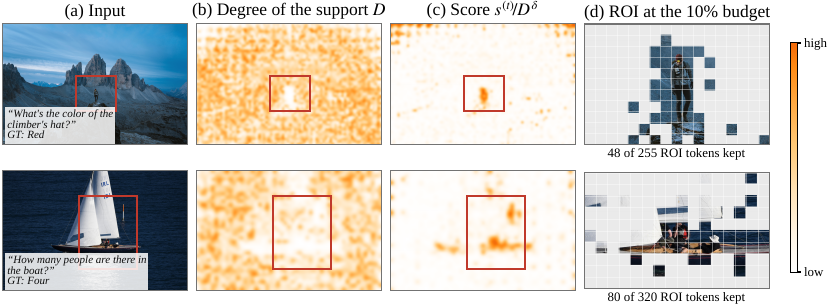}
\caption{\textbf{Local score--degree relationships.} HR-Bench 8K examples at the $10\%$ retention budget. Degree and corrected score are negatively associated in the displayed cases (reported correlations $-0.854$ and $-0.931$). Retained regions are shown for visual comparison. The existing visualization uses $t=5$, $\delta=1$; the default component table uses $t=3$. Heatmaps are interpolated and scaled separately.}
\label{fig:sheafprune_mechanism}
\end{figure}

\textbf{What the operator computes.} Figure~\ref{fig:sheafprune_mechanism} links the score-based analysis to two HR-Bench 8K examples. Regions with lower neighborhood degree receive higher corrected scores in these samples, with reported degree--score correlations of $-0.854$ and $-0.931$. This is consistent with the explicit division by $d_i^\delta$, but propagated mass also enters the score: the figure is not evidence that low degree alone determines selection. The displayed retained regions illustrate the combined operator, without isolating the effect of question conditioning.

\textbf{Graph construction.} The existing $k$-NN comparison reports a $23$--$24$-point decrease in the DocVQA score for tested $k\in\{16,32,64,128\}$ (Appendix~\ref{app:hyper}). A sample-wise threshold permits variable neighborhood sizes while preserving symmetric support. Replacing it by $k$-NN changes neighborhood structure and potentially symmetry and degree distribution together, so the comparison motivates examining graph construction but does not isolate one cause.

\Needspace{10\baselineskip}
\textbf{Where selection stops paying off.} Mapping retained tokens back to the image grid reveals a limitation of score-based selection: neither propagation nor degree correction explicitly requires the selected subset to cover the image. At $p=0.9$, GQA grid coverage is $0.417$ for SheafPrune against $0.500$ for random selection. Absolute counts also matter: $345$ candidates leave about $35$ tokens, whereas $4096$ HR-Bench candidates leave about $410$, with observed coverage $0.844$. These examples distinguish a retention ratio from the amount and distribution of surviving evidence. They do not establish the cause of the LLaVA-1.5 or InternVL3 exceptions, because task content, tokenization and absolute counts differ simultaneously. Coverage-aware selection is a direction for testing, rather than an already validated remedy. Appendix~\ref{app:coverage} provides the full coverage measurements.

\textbf{Sensitivity within the tested setting.} The existing Qwen2.5-VL sweep at $p=0.9$ gives four-benchmark mean scores of $53.85$--$54.23$ for $t\in\{1,3,5,10\}$ and $53.06$ for $t=20$. Thus, additional propagation is not uniformly beneficial in this sweep. The separate absolute-similarity-threshold experiment has its largest displayed mean at $s=0.5$; it should not be conflated with the default quantile parameter $q_v$. These measurements describe the tested settings and do not supply a noise threshold or a universal optimal depth (Appendix~\ref{app:hyper}).

\section{Conclusion}
\label{sec:conclusion}
SheafPrune uses sheaf-guided cross-modal comparison, support-constrained visual propagation and degree correction to rank visual tokens under a fixed budget. One trained checkpoint serves all evaluated retention ratios. The fixed-configuration ablations show that connectivity and the ranking operator preserve much of the full method's performance even from uniform initialization, while degree-only ranking is insufficient. Under the reported best-$\delta$ protocol, SheafPrune leads $22$ of $28$ backbone--budget settings; on LLaVA-NeXT, retaining $10\%$ of candidates gives $90.42\%$ RA, $5.15\times$ net prefill speedup and $84.0\%$ end-to-end FLOPs reduction. These results characterize the measured accuracy--prefill trade-off across four backbones rather than a tuning-free deployment curve. Coverage observations further distinguish score quality from spatial coverage. Overall, SheafPrune advances graph-based visual token pruning through a sheaf-guided framework for budget-constrained selection, providing a methodological and empirical foundation for future multimodal compression methods that preserve useful visual evidence while reducing inference cost.

\subsection*{AI use statement}

In this work, we used generative AI tools, including Anthropic
Claude Code with an Opus model and OpenAI Codex, to assist with
code implementation, code review and testing, information
retrieval, and critical feedback on the manuscript. The research
ideas, algorithmic derivations, experimental design, and research
plan were developed by the authors.

Claude Code assisted in implementing our designed functions
and algorithmic components, including those involving sheaf
theory and graph diffusion. AI tools also assisted with code
review and testing to help identify potential bugs and gaps in
test coverage. Codex assisted in finding related papers for
comparison, locating datasets and model download sources, and
troubleshooting network and system issues. We also used
generative AI to provide critical feedback on the manuscript
and help identify potential weaknesses in its reasoning and
presentation. 

We manually reviewed the AI-assisted code, tracing
each function's logic from its inputs to its outputs and
checking the algorithmic implementation against the mathematical
formulations. All code was executed directly by the authors.
Each paper identified through AI-assisted search was individually
checked using search engines to verify its existence and guard
against fabricated references.

We take responsibility for the final content of this work,
including text, claims, and artifacts produced with the aid of
generative AI.
\subsection*{Ethics statement}

Our experiments use the publicly available LLaVA-NeXT-Data
dataset\citep{liu2024llavanext}, which combines visual instruction-following data
with document, chart, and diagram understanding data.
According to the dataset documentation, the original dataset
creators filtered samples with potential privacy concerns or
harmful content from the 15K visual instruction subset
collected through the LLaVA demo. We acknowledge that these
upstream filtering measures do not eliminate all potential
risks and that the dataset may retain biases, inaccuracies,
or sensitive content inherited from its constituent sources.
These limitations should be considered when interpreting
the experimental results and assessing potential applications
of the resulting models.

\subsection*{Reproducibility statement}

The formulation and algorithmic procedure of SheafPrune are
presented in Section~\ref{sec:method}. Implementation settings
and training configurations are described in
Appendices~\ref{app:implementation} and~\ref{app:training_setup}.
The benchmark splits, evaluation metrics, generation settings,
and efficiency measurement protocol are documented in
Appendix~\ref{app:eval_protocol}, while baseline implementations
and token-budget accounting are described in
Appendix~\ref{app:baselines}. The reporting protocol distinguishes
per-point degree-exponent selection in the main comparisons
from the fixed default configuration used for component
ablations. Full reported benchmark scores and component
ablation results are provided in
Appendices~\ref{app:rawscores} and~\ref{app:component_ablation}.
Our code is included in the supplementary material. Model weights and checkpoints are omitted due to file-size limits. We will release the full source code and the trained selector checkpoints, including all weights, after the double-blind review.

\bibliographystyle{iclr2027_conference}
\bibliography{iclr2027_conference}

\appendix

\section{Sheaf-Theoretic Background and Method Positioning}
\label{app:sheaf_background}
The sheaf perspective guides heterogeneous comparison and relation-constrained transfer. A cellular sheaf on a graph assigns vector spaces to vertices and edges and linear maps from each incident vertex space to the edge space \citep{hansen2019spectral}. For an oriented edge $e=(u,v)$,
\begin{equation}
R_{v\rightarrow e}:\mathcal F(v)\rightarrow\mathcal F(e),\qquad
(\delta_{\mathcal F}y)_e=R_{u\rightarrow e}y_u-R_{v\rightarrow e}y_v.
\label{eq:app_restriction_map}
\end{equation}
The squared edge disagreement and its aggregate are
\begin{align}
\mathcal E_e(y)&=\|R_{u\rightarrow e}y_u-R_{v\rightarrow e}y_v\|_2^2,\label{eq:app_local_energy}\\
\mathcal E_{\mathcal F}(y)&=\tfrac12\sum_e\mathcal E_e(y)
=\tfrac12\|\delta_{\mathcal F}y\|_2^2
=\tfrac12y^\top L_{\mathcal F}y,\qquad
L_{\mathcal F}=\delta_{\mathcal F}^\top\delta_{\mathcal F}.\label{eq:app_sheaf_energy}
\end{align}
An associated linear diffusion has the form
\begin{equation}
y^{(\ell+1)}=(I-\eta L_{\mathcal F})y^{(\ell)}.
\label{eq:app_sheaf_diffusion}
\end{equation}
This equation provides background, not the visual update used by SheafPrune.

\textbf{Conditional correspondence for cross-modal comparison.}
Fix a sample, its text--vision edge set and its encoded, normalized local states
$y_j=\operatorname{LN}_T(\widetilde x_j^t)$ and $y_i=\operatorname{LN}_V(x_i^v)$.
Give each vertex the selector-interface space $\mathbb R^d$ and each edge $\mathbb R^h$.
For that sample define the positive scalars
$a_j=\|F_Ty_j\|_2+\epsilon$ and $a_i=\|F_Vy_i\|_2+\epsilon$, and set
\begin{equation}
R_{j\rightarrow e}=F_T/a_j,\qquad R_{i\rightarrow e}=F_V/a_i,
\qquad E_{ji}=\|R_{j\rightarrow e}y_j-R_{i\rightarrow e}y_i\|_2^2.
\label{eq:app_conditional_sheaf}
\end{equation}
Once the sample-dependent scalars are fixed, these are linear maps on a graph and specify a cellular sheaf. The evaluated edge energy is exactly the stabilized projected disagreement. The overall map from raw inputs to energy remains nonlinear because encoding, LayerNorm, normalization and edge selection depend on the input. This conditional correspondence does not turn the full selector into a fixed linear sheaf operator.

\textbf{What this interpretation contributes.}
The correspondence makes the design choice explicit: text and visual states are compared after type-specific transformations into an edge space. With exact unit normalization, the same energy is also a cosine-based comparison of projected representations. This equivalence is acknowledged rather than treated as a separate source of expressivity. Unlike Neural Sheaf Diffusion \citep{bodnar2022neural}, the visual stage here uses the independently constructed scalar transition in Eq.~\eqref{eq:method_transition}; it does not propagate with the cross-modal $L_{\mathcal F}$. Neighborhood-count correction is a further ranking operation. Sheaf guidance is therefore a design rationale with a precise cross-modal correspondence, while the empirical contribution is assessed through token-selection performance and operator ablations.

\section{Notation, Masks, and Fixed Budget}
\label{app:notation_masks}

The masks of Section~\ref{sec:method_overview} are formed per sample: for sample $b$ in a batch, with sequences $X_T^{(b)}\in\mathbb R^{N_t\times d}$ and $X_V^{(b)}\in\mathbb R^{N_v\times d}$, the textual validity mask $\bm u\in\{0,1\}^{N_t}$, the visual validity mask $\bm v\in\{0,1\}^{N_v}$ and the candidate mask $\bm c\in\{0,1\}^{N_v}$ with $c_i\leq v_i$ handle variable-length text, AnyRes visual sequences and non-prunable structural positions such as \texttt{image\_newline} ($v_i=1$, $c_i=0$). Given a target retention ratio $\rho\in(0,1]$, the candidate set and retention budget of sample $b$ are
\begin{equation}
    \mathcal C^{(b)}
    =
    \left\{i:c_i^{(b)}=1\right\},
    \qquad
    k^{(b)}
    =
    \min\!\left\{
    \left|\mathcal C^{(b)}\right|,
    \max\!\left(
    1,
    \left\lceil\rho\left|\mathcal C^{(b)}\right|\right\rceil
    \right)
    \right\}.
    \label{eq:app_budget}
\end{equation}
Because the number of candidates may differ across samples, $k^{(b)}$ is computed independently for each sample. Every softmax, Top-$K$ operation, degree statistic, min--max normalization, z-score, and final ranking is evaluated only on its valid support.

\section{Leakage-Free, Budget-Conditioned Text Encoding}
\label{app:text_encoder}

The selector receives only the system prompt, dialogue history completed before the target turn, and the current question. The current target answer, content following the target turn, expanded image-placeholder positions, padding, and configured control tokens are excluded. For each multi-turn training sample, a target turn is first selected and the textual validity mask $\bm u$ is then constructed accordingly. The selector therefore has the same information boundary at training and inference and cannot exploit answer tokens to form a non-deployable pruning policy.

Because the frozen language-model input embeddings do not themselves contain explicit token order, we introduce a lightweight bidirectional encoder outside the language model. Its complete computation is
\begin{align}
    H_T^{(0)}
    &=
    W_{\mathrm{down}}
    \operatorname{LN}_{\mathrm{in}}(X_T)
    +P+\bm 1\,g(\rho)^\top,
    \label{eq:app_text_input}\\
    g(\rho)
    &=
    \operatorname{MLP}\!\left([\rho,\log\rho]\right),
    \label{eq:app_budget_embedding}\\
    H_T
    &=
    \operatorname{Enc}_{\mathrm{bi}}
    \left(H_T^{(0)};\bm u\right),
    \qquad
    \widetilde X_T=W_{\mathrm{up}}H_T.
    \label{eq:app_text_output}
\end{align}
Here, $W_{\mathrm{down}}\in\mathbb R^{d_e\times d}$ and $W_{\mathrm{up}}\in\mathbb R^{d\times d_e}$ form a bottleneck projection, $P\in\mathbb R^{N_t\times d_e}$ is a learned absolute positional embedding, and $g(\rho)\in\mathbb R^{d_e}$ is the retention-ratio condition. The bidirectional encoder $\operatorname{Enc}_{\mathrm{bi}}$ uses $\bm u$ as its padding mask and is independent of the causal attention layers in the frozen language model. We use a bottleneck dimension of $d_e=256$, two bidirectional attention layers, four attention heads, and a maximum selector text length of 1024.

\section{Cross-Modal Compatibility and Text-to-Vision Propagation}
\label{app:cross_modal}

Textual and visual tokens are projected into a shared $h$-dimensional space after modality-specific input normalization:
\begin{align}
    z_j^t
    &=F_T\operatorname{LN}_T(\widetilde x_j^t),
    &
    \widehat z_j^t
    &=\frac{z_j^t}{\lVert z_j^t\rVert_2+\epsilon},
    \label{eq:app_text_map}\\
    z_i^v
    &=F_V\operatorname{LN}_V(x_i^v),
    &
    \widehat z_i^v
    &=\frac{z_i^v}{\lVert z_i^v\rVert_2+\epsilon}.
    \label{eq:app_visual_map}
\end{align}
The maps satisfy $F_T,F_V\in\mathbb R^{h\times d}$, with $h=128$ by default. Post-projection $\ell_2$ normalization bounds the cross-modal edge energy:
\begin{equation}
    E_{ji}
    =
    \left\|\widehat z_j^t-\widehat z_i^v\right\|_2^2
    =
    \|\widehat z_j^t\|_2^2+\|\widehat z_i^v\|_2^2
    -2\left\langle\widehat z_j^t,\widehat z_i^v\right\rangle
    \in[0,4].
    \label{eq:app_edge_energy}
\end{equation}
For exact unit vectors this reduces to $2-2\langle\widehat z_j^t,\widehat z_i^v\rangle$. With the stated additive $\epsilon$, the squared norms are at most one and the full expansion above is the exact identity, including zero projections. The edge energy is converted into affinity by a negative-exponential kernel:
\begin{equation}
    A_{ji}
    =
    u_jv_i
    \exp\!\left(-\frac{E_{ji}}{\tau_E}\right),
    \qquad
    \tau_E=\exp(\ell_\tau)>0.
    \label{eq:app_affinity}
\end{equation}
The temperature $\tau_E$ is initialized once from the median valid edge energy in the first minibatch and is subsequently optimized as an ordinary learnable parameter. Textual importance is defined by
\begin{equation}
    a_j=w_{\mathrm{imp}}^\top\widehat z_j^t+b_{\mathrm{imp}},
    \qquad
    \alpha_j
    =
    \frac{u_j\exp(a_j)}{\sum_{j'}u_{j'}\exp(a_{j'})}.
    \label{eq:app_text_importance}
\end{equation}
In the default path, $\alpha$ is used only as the textual propagation source and is not inserted into the exponential kernel in Eq.~\eqref{eq:app_affinity}. This separation makes $A_{ji}$ describe cross-modal geometric compatibility and $\alpha_j$ describe the contribution of textual token $j$ to the current task.

For each valid textual token, only the $K$ valid visual neighbors with the highest affinity are retained. The neighborhood size adapts to the visual sequence length:
\begin{equation}
    K
    =
    \min\!\left\{
    N_v^{\mathrm{valid}},
    \max\!\left(
    K_{\min},
    \left\lceil q_{\mathrm{tv}}N_v^{\mathrm{valid}}\right\rceil
    \right)
    \right\}.
    \label{eq:app_bipartite_k}
\end{equation}
The default values are $K_{\min}=32$ and $q_{\mathrm{tv}}=0.05$. Invalid visual columns are masked before row-wise Top-$K$ so that padding cannot occupy a valid neighbor slot. Let $M$ denote the sparse support and $\widetilde A=A\odot M$. Bipartite propagation is then
\begin{align}
    D_T
    &=\operatorname{diag}(\widetilde A\bm 1)+\epsilon I,
    &
    D_V
    &=\operatorname{diag}(\widetilde A^\top\bm 1)+\epsilon I,
    \label{eq:app_bipartite_degrees}\\
    P
    &=D_T^{-1/2}\widetilde A D_V^{-1/2},
    &
    r
    &=P^\top\alpha.
    \label{eq:app_bipartite_propagation}
\end{align}
The relevance $r$ is min--max normalized only over valid visual positions to obtain $\widehat r$. If $\max(r)-\min(r)<10^{-6}$ over valid positions, we set $\widehat r_i=0.5$ at all valid positions instead of amplifying numerical noise into an unstable ranking.

\section{Visual Graph, Constrained Propagation, and Degree Correction}
\label{app:visual_graph}

The visual graph is independent of the text--vision bipartite graph. For each pair of valid off-diagonal visual tokens, we first compute
\begin{equation}
    \operatorname{sim}_{ii'}
    =
    \cos(x_i^v,x_{i'}^v).
    \label{eq:app_visual_similarity}
\end{equation}
Let $\xi_{q_v}$ be the $(1-q_v)$ quantile of the valid off-diagonal similarities within a sample. The neighborhood of node $i$ is
\begin{equation}
    \mathcal N(i)
    =
    \{i\}
    \cup
    \left\{
    i'\neq i:
    v_{i'}=1,
    \operatorname{sim}_{ii'}\geq\xi_{q_v}
    \right\}.
    \label{eq:app_visual_support}
\end{equation}
Self-loops are inserted explicitly and are excluded from quantile estimation, so $q_v$ denotes the target retained proportion among valid off-diagonal relations. A sample-wise global threshold preserves a symmetric support for the undirected similarity graph. We use $q_v=0.05$ by default and do not center the features before computing cosine similarity.

The transition matrix restricts propagation to the selected neighbors $\mathcal N(i)$:
\begin{equation}
    T_{ii'}
    =
    \begin{cases}
    \displaystyle
    \frac{\exp(\operatorname{sim}_{ii'})}
    {\sum_{l\in\mathcal N(i)}\exp(\operatorname{sim}_{il})},
    &i'\in\mathcal N(i),\\[10pt]
    0,
    &i'\notin\mathcal N(i).
    \end{cases}
    \label{eq:app_masked_transition}
\end{equation}
Edge existence is determined by the support in Eq.~\eqref{eq:app_visual_support}, not by whether the similarity value is positive. A retained edge with negative cosine similarity therefore still receives nonzero propagation weight, while every non-edge receives exactly zero weight.

The visual-norm prior is $n_i=\lVert x_i^v\rVert_2$. Let $\overline n$ denote its min--max normalization over valid visual positions. To avoid an empirical scale mismatch between the two branches, we standardize $\widehat r$ over the candidate set before early fusion:
\begin{align}
    \widetilde r_i
    &=
    \frac{
    \widehat r_i-\operatorname{mean}_{\mathcal C}(\widehat r)
    }{
    \operatorname{std}_{\mathcal C}(\widehat r)+\epsilon
    },
    \label{eq:app_relevance_zscore}\\
    s_i^{(0)}
    &=
    (1-\lambda)\overline n_i
    +
    \lambda\,
    \operatorname{std}_{\mathcal C}(\overline n)\,
    \widetilde r_i,
    \qquad
    \lambda=\sigma(\ell_\lambda).
    \label{eq:app_early_fusion}
\end{align}
Treating $s^{(\ell)}$ as a row vector, we perform propagation and degree correction as follows:
\begin{equation}
    s^{(\ell+1)}=s^{(\ell)}T,
    \quad \ell=0,\ldots,t-1,
    \qquad
    f_i=\frac{s_i^{(t)}}{d_i^\delta},
    \qquad
    d_i=|\mathcal N(i)|.
    \label{eq:app_diffusion_degree}
\end{equation}
Defaults are $t=3$ and $\delta=1$. Degree correction uses neighborhood counts, not weighted row sums or weight signs, and is distinct from transition-weight normalization.

\subsection*{Finite-step redistribution and long-run degree bias}
Let $W_{ij}=\mathbb1[j\in\mathcal N(i)]\exp(\operatorname{sim}_{ij})$, $w_i=\sum_jW_{ij}$ and $D_w=\operatorname{diag}(w_i)$. On valid visual nodes the symmetric similarity and global threshold give symmetric $W$, positive self-loops and $T=D_w^{-1}W$. For uniform scores, Eq.~\eqref{eq:method_uniform_step} follows by matrix multiplication: column sums, not row sums, determine the received score. Equal row sums therefore do not force equal rankings.

\textbf{Long-run statement.} For a connected component $C$, let $m_C=\sum_{j\in C}s_j^{(0)}$. Since each component is finite, connected and has positive self-loops, its random walk is irreducible and aperiodic. Its stationary distribution is $\pi_i=w_i/\sum_{j\in C}w_j$, and hence
\begin{equation}
\lim_{t\to\infty}s_i^{(t)}=m_C\frac{w_i}{\sum_{j\in C}w_j},\qquad
\lim_{t\to\infty}f_i=\frac{m_C}{\sum_{j\in C}w_j}\frac{w_i}{d_i^\delta},\quad i\in C.
\label{eq:app_stationary_ranking}
\end{equation}
\emph{Proof.} Symmetry gives detailed balance, $\pi_iT_{ij}=W_{ij}/\sum_{j\in C}w_j=\pi_jT_{ji}$. Irreducibility and aperiodicity imply $T_C^t\to\bm1\pi^\top$. Left-multiplication by the initial row vector gives the first limit; division by $d_i^\delta$ gives the second. This linear statement also holds for signed initial scores. If $m_C=0$ the limiting component vanishes; if $m_C<0$, its ranking direction is reversed.

For $\delta=1$ and positive component mass, the limiting ranking within that component depends on $w_i/d_i$, the mean retained exponential similarity, rather than on neighborhood count alone. Count correction therefore changes degree dependence but need not eliminate all connectivity effects. Components may carry different total masses. The actual method uses finite $t=3$ by default: these limits do not establish convergence within three steps, question independence, or an oversmoothing explanation for the measured accuracy losses. The finite-step identity and the ablations are the direct evidence used in the main text.

\section{Fixed-Budget Selection and End-to-End Training}
\label{app:training}

The final score is first standardized over the candidate set:
\begin{equation}
    \overline f_i
    =
    \frac{
    f_i-\operatorname{mean}_{\mathcal C}(f)
    }{
    \operatorname{std}_{\mathcal C}(f)+\epsilon
    }.
    \label{eq:app_final_zscore}
\end{equation}
Let $\theta_k$ be the $k$-th largest standardized candidate score. The hard mask, smooth surrogate, and straight-through estimator are
\begin{align}
    m_i^{\mathrm{hard}}
    &=\mathbb 1[i\in\operatorname{TopK}_{\mathcal C}(\overline f;k)],
    \label{eq:app_hard_mask}\\
    m_i^{\mathrm{soft}}
    &=\sigma\!\left(
    \frac{\overline f_i-\operatorname{sg}(\theta_k)}{\beta}
    \right),
    \label{eq:app_soft_mask}\\
    m_i
    &=m_i^{\mathrm{soft}}
    +\operatorname{sg}\!\left(
    m_i^{\mathrm{hard}}-m_i^{\mathrm{soft}}
    \right),
    \label{eq:app_ste_mask}
\end{align}
where $\operatorname{sg}$ denotes stop-gradient. The forward pass uses the exact hard Top-$k$ set, while the backward pass uses the sigmoid surrogate to pass gradients near the selection boundary. Exact ties are resolved using a fixed order of original token indices to ensure reproducibility. Every non-candidate structural token is assigned $m_i=1$.

During training, visual gating is applied to the key dimension of every language-model attention layer. Let $p_{q\kappa}$ be the original attention probability from query position $q$ to key position $\kappa$, and let $\mu_\kappa$ be the corresponding gate. The gated probability is
\begin{equation}
    p'_{q\kappa}
    =
    \frac{
    \mu_\kappa p_{q\kappa}
    }{
    \sum_{\kappa'}\mu_{\kappa'}p_{q\kappa'}+\epsilon
    }.
    \label{eq:app_multiplicative_gate}
\end{equation}
Visual positions use their corresponding $m_i$, whereas textual and mandatory structural positions use $\mu_\kappa=1$. Because at least the textual positions remain unpruned, the denominator of Eq.~\eqref{eq:app_multiplicative_gate} remains positive.

Training uses only standard next-token cross-entropy at the target-answer positions:
\begin{equation}
    \mathcal L
    =
    \mathcal L_{\mathrm{CE}}
    =
    -\frac{1}{|\mathcal A|}
    \sum_{q\in\mathcal A}
    \log p_\theta(y_q\mid y_{<q},X_T,X_V;m),
    \label{eq:app_training_loss}
\end{equation}
where $\mathcal A$ is the set of current target-answer positions. The vision encoder, multimodal projector, and language model remain frozen, and only SheafPrune is updated. Gradients reach the selector through the frozen generation path and multiplicative gating. Here, end-to-end training means supervising SheafPrune directly with the final language-generation objective while keeping the LLaVA-NeXT backbone frozen.

\section{Inference-Time Physical Removal and Position Handling}
\label{app:physical_pruning}

Training-time gating provides gradients to the selector. At inference, unselected visual tokens are physically removed before entering the language model. Let $\mathcal I_{\mathrm{keep}}$ be the ordered index set containing all hard-selected candidates and mandatory structural positions. We compute
\begin{equation}
    X_V^{\mathrm{keep}}
    =
    \operatorname{Gather}(X_V,\mathcal I_{\mathrm{keep}}),
    \qquad
    \operatorname{pos}^{\mathrm{keep}}
    =
    \operatorname{Gather}(
    \operatorname{pos}^{\mathrm{orig}},
    \mathcal I_{\mathrm{keep}}).
    \label{eq:app_physical_gather}
\end{equation}
The retained tokens preserve their original position indices instead of receiving newly generated contiguous \texttt{arange} indices, thereby maintaining the positional relations used by rotary embeddings. The attention mask, visual tokens, and position indices must be gathered with the same index set.

Training-time gating and inference-time physical removal use the same hard subset but remain different execution paths. In the current fp32 validation, the maximum logit difference is $5.147\times10^{-3}$, which remains above the strict $10^{-3}$ equivalence threshold. We therefore do not claim strict numerical equivalence between the two paths. All reported accuracy and efficiency measurements use the physical-removal path. Algorithm~\ref{alg:app_sheafprune} instantiates Algorithm~\ref{alg:method_sheafprune} with every appendix equation, default value, and support restriction made explicit.

\begin{algorithm}[!htbp]
    \caption{Single-sample SheafPrune, with the defining equation of every step. Superscripts $(b)$ are omitted; each statistic is evaluated only on its valid support.}
    \label{alg:app_sheafprune}
    \small
    \begin{algorithmic}[1]
        \Require $X_T\in\mathbb R^{N_t\times d}$, $X_V\in\mathbb R^{N_v\times d}$; $\bm u,\bm v,\bm c$; $\rho$; defaults $h,K_{\min},q_{\mathrm{tv}},q_v,t,\delta,\beta$
        \Ensure $m\in\{0,1\}^{N_v}$, ordered index set $\mathcal I_{\mathrm{keep}}$
        \Statex \textbf{Stage 1: budget and textual context}
        \State $\mathcal C\gets\{i:c_i=1\}$,\quad
               $k\gets\min\{|\mathcal C|,\max(1,\lceil\rho|\mathcal C|\rceil)\}$
               \Comment{Eq.~\eqref{eq:app_budget}}
        \State $\widetilde X_T\gets W_{\mathrm{up}}\operatorname{Enc}_{\mathrm{bi}}
               \!\left(W_{\mathrm{down}}\operatorname{LN}_{\mathrm{in}}(X_T)+P+\bm 1g(\rho)^\top;\bm u\right)$
               \Comment{Eq.~\eqref{eq:app_text_input}--\eqref{eq:app_text_output}}
        \Statex \textbf{Stage 2: cross-modal relevance}
        \State $\widehat z_j^t\gets F_T\operatorname{LN}_T(\widetilde x_j^t)/(\lVert\cdot\rVert_2+\epsilon)$,\quad
               $\widehat z_i^v\gets F_V\operatorname{LN}_V(x_i^v)/(\lVert\cdot\rVert_2+\epsilon)$
               \Comment{Eq.~\eqref{eq:app_text_map},~\eqref{eq:app_visual_map}}
        \State $E_{ji}\gets\|\widehat z_j^t-\widehat z_i^v\|_2^2\in[0,4]$,\quad
               $A_{ji}\gets u_jv_i\exp(-E_{ji}/\tau_E)$
               \Comment{Eq.~\eqref{eq:app_edge_energy},~\eqref{eq:app_affinity}}
        \State $\alpha_j\gets u_j\exp(a_j)/\textstyle\sum_{j'}u_{j'}\exp(a_{j'})$,\quad
               $a_j=w_{\mathrm{imp}}^\top\widehat z_j^t+b_{\mathrm{imp}}$
               \Comment{Eq.~\eqref{eq:app_text_importance}}
        \State $K\gets\min\{N_v^{\mathrm{valid}},\max(K_{\min},\lceil q_{\mathrm{tv}}N_v^{\mathrm{valid}}\rceil)\}$,\quad
               $\widetilde A\gets A\odot M$
               \Comment{Eq.~\eqref{eq:app_bipartite_k}}
        \State $P\gets D_T^{-1/2}\widetilde AD_V^{-1/2}$,\quad
               $\widehat r\gets\operatorname{MinMax}_{\bm v}(P^\top\alpha)$
               \Comment{Eq.~\eqref{eq:app_bipartite_degrees},~\eqref{eq:app_bipartite_propagation}}
        \Statex \textbf{Stage 3: structure-constrained propagation}
        \State $\xi_{q_v}\gets\operatorname{Quantile}_{1-q_v}(\operatorname{sim}_{i\neq i'})$,\quad
               $\mathcal N(i)\gets\{i\}\cup\{i':\operatorname{sim}_{ii'}\geq\xi_{q_v}\}$
               \Comment{Eq.~\eqref{eq:app_visual_support}}
        \State $T_{ii'}\gets\mathbb 1[i'\in\mathcal N(i)]\exp(\operatorname{sim}_{ii'})/
               \textstyle\sum_{l\in\mathcal N(i)}\exp(\operatorname{sim}_{il})$
               \Comment{Eq.~\eqref{eq:app_masked_transition}}
        \State $s^{(0)}\gets(1-\lambda)\overline n+\lambda\operatorname{std}_{\mathcal C}(\overline n)\,
               \operatorname{zscore}_{\mathcal C}(\widehat r)$,\quad $\lambda=\sigma(\ell_\lambda)$
               \Comment{Eq.~\eqref{eq:app_relevance_zscore},~\eqref{eq:app_early_fusion}}
        \For{$\ell=0,\ldots,t-1$}
            \State $s^{(\ell+1)}\gets s^{(\ell)}T$
                   \Comment{Eq.~\eqref{eq:app_diffusion_degree}}
        \EndFor
        \State $f_i\gets s_i^{(t)}/d_i^{\delta}$,\quad
               $\overline f\gets\operatorname{zscore}_{\mathcal C}(f)$
               \Comment{Eq.~\eqref{eq:app_diffusion_degree},~\eqref{eq:app_final_zscore}}
        \Statex \textbf{Stage 4: fixed-budget selection}
        \State $m_i^{\mathrm{hard}}\gets\mathbb 1[i\in\operatorname{TopK}_{\mathcal C}(\overline f;k)]$,\quad
               $m_i^{\mathrm{soft}}\gets\sigma\!\left((\overline f_i-\operatorname{sg}\theta_k)/\beta\right)$
               \Comment{Eq.~\eqref{eq:app_hard_mask},~\eqref{eq:app_soft_mask}}
        \State $m_i\gets m_i^{\mathrm{soft}}+\operatorname{sg}(m_i^{\mathrm{hard}}-m_i^{\mathrm{soft}})$,\quad
               $m_i\gets1$ if $c_i=0,v_i=1$
               \Comment{Eq.~\eqref{eq:app_ste_mask}}
        \State \textbf{train:} $p'_{q\kappa}\gets\mu_\kappa p_{q\kappa}/(\sum_{\kappa'}\mu_{\kappa'}p_{q\kappa'}+\epsilon)$
               \Comment{Eq.~\eqref{eq:app_multiplicative_gate}}
        \State \textbf{infer:} $\mathcal I_{\mathrm{keep}}\gets\{i:m_i=1\}$,\quad
               $(X_V,\operatorname{pos})\gets\operatorname{Gather}(\cdot,\mathcal I_{\mathrm{keep}})$
               \Comment{Eq.~\eqref{eq:app_physical_gather}}
        \State \Return $m,\mathcal I_{\mathrm{keep}}$
    \end{algorithmic}
\end{algorithm}

\section{Degenerate Cases and Numerical Stability}
\label{app:edge_cases}

If a sample contains no visual tokens, the selector skips graph construction and follows the text-only path. If no prunable candidate exists, all valid structural tokens are retained. A sample with no valid textual token violates the current multimodal-input contract and is skipped and recorded. If $\rho=1$ or $k=|\mathcal C|$, all candidate gates equal one, which also provides the unpruned path used for validation. If the variance of the candidate scores is nearly zero, ties are resolved deterministically by original token index so that different computational backends return the same subset.

Cross-modal energy, affinity, visual similarity, graph construction, and propagation are computed in fp32 to reduce underflow and ranking errors in large exponential kernels and softmax operations. After the selection mask is produced, it is converted to the hidden dtype used by the language model. Invalid positions are masked before normalization rather than multiplied by zero afterward; otherwise, invalid positions could first receive probability mass and alter the relative weights among valid positions.

\section{Implementation Configuration}
\label{app:implementation}

Table~\ref{tab:app_hyperparameters} summarizes the default method configuration. Explicit retention-ratio conditioning allows one SheafPrune checkpoint to support multiple visual-token budgets. The bidirectional text encoder, modality-specific maps, textual-importance head, retention conditioner, and learnable scalar parameters introduce approximately $4.5$M trainable parameters in total. The $d=4096$ entry describes the training interface; it is not a specification of each evaluated backbone's native embedding width.

\begin{table}[!htbp]
    \caption{Default SheafPrune configuration.}
    \label{tab:app_hyperparameters}
    \centering
    \small
    \begin{tabular}{@{}llc@{\qquad}llc@{}}
        \toprule
        Component & Symbol & Default & Component & Symbol & Default\\
        \midrule
        training-interface dimension & $d$ & $4096$ & visual-graph relation ratio & $q_v$ & $0.05$\\
        text bottleneck & $d_e$ & $256$ & propagation steps & $t$ & $3$\\
        text-encoder layers & $L_e$ & $2$ & degree exponent & $\delta$ & $1$\\
        text-encoder heads & -- & $4$ & fusion mode & -- & early\\
        max selector text length & -- & $1024$ & visual cosine centering & -- & false\\
        comparison dimension & $h$ & $128$ & propagation normalization & -- & masked\\
        min bipartite neighborhood & $K_{\min}$ & $32$ & STE temperature & $\beta$ & $2.0\!\to\!0.1$\\
        bipartite ratio & $q_{\mathrm{tv}}$ & $0.05$ & optimizer & -- & AdamW\\
        graph precision & -- & fp32 & initial learning rate & -- & $2\times10^{-4}$\\
        trainable parameters & -- & $\approx4.5$M & & & \\
        \bottomrule
    \end{tabular}
\end{table}

\section{Auxiliary Space and Speed Condition}
\label{app:efficiency}

When the textual self-attention matrix, text--vision affinity matrix, and vision--vision similarity matrix are directly materialized, the peak auxiliary space of the selector, excluding input and output representations and model parameters, is
\begin{equation}
    M_{\mathrm{SP}}^{\mathrm{aux}}
    =
    \mathcal O\!\left(
    N_t^2+N_tN_v+N_v^2
    \right).
    \label{eq:app_aux_memory}
\end{equation}
When $N_v\gg N_t$, the complete visual-similarity matrix is typically the dominant relational workspace. Sparsification lowers the practical storage constant during propagation, but it does not remove the $N_v^2$ term while the complete similarity matrix is explicitly constructed. Further reduction of peak space requires blockwise similarity computation or neighbor construction that does not materialize all pairwise relations.

Let $C_{\mathrm{sel}}^{(a)}$ be the selection cost of method $a$, $n_\ell^{(a)}$ the sequence length processed by language-model layer $\ell$, and $C_\ell(n)$ the cost of processing length $n$ at that layer. Selecting once before the language model gives a constant length across layers, and SheafPrune is faster than method $a$ whenever
\begin{align}
    n_\ell^{(\mathrm{SP})}
    &=N_t+N_v^{\mathrm{keep}},
    \qquad
    N_v^{\mathrm{keep}}=k+N_{\mathrm{str}},
    \qquad
    \forall\ell\in\{1,\ldots,L\},
    \label{eq:method_kept_length}\\
    C_{\mathrm{sel}}^{(\mathrm{SP})}-C_{\mathrm{sel}}^{(a)}
    &<
    \sum_{\ell=1}^{L}
    \left[
    C_\ell\!\left(n_\ell^{(a)}\right)
    -
    C_\ell\!\left(N_t+N_v^{\mathrm{keep}}\right)
    \right].
    \label{eq:method_speed_condition}
\end{align}
Latency therefore depends jointly on selector overhead and the accumulated language-model computation saved by earlier sequence shortening. SheafPrune determines the subset once before the decoder, allowing the compressed sequence to be used from the first layer and avoiding layer-wise rescoring, token recycling, or online calibration. For methods that also perform one-shot physical removal before the decoder, the downstream language-model cost is approximately matched under the same budget, and the speed difference is governed primarily by the actual kernels, memory access, dynamic branches, and synchronization in the selector. We therefore report the speed advantage over the compared training-free methods as an empirical system result under a unified hardware and implementation protocol, rather than as a universal consequence of asymptotic complexity. The selector's own training is a one-off offline cost that does not enter the inference path, and the reported latencies include its forward cost.

The measured total includes the selector's forward cost. A full stage-level breakdown would additionally separate text encoding, cross-modal scoring, graph construction, propagation and Top-$k$; no such breakdown is inferred from the aggregate values. The vision encoder and multimodal projector execute before pruning, so their computation is not reduced by token removal.

\section{Evaluation Protocol}
\label{app:eval_protocol}

\textbf{Benchmark sources.}
We use ChartQA \citep{masry2022chartqa}, DocVQA \citep{mathew2021docvqa}, TextVQA \citep{singh2019textvqa}, HR-Bench 4K/8K \citep{wang2025hrbench}, GQA \citep{hudson2019gqa}, VQAv2 \citep{goyal2017vqav2}, MMBench-EN \citep{liu2024mmbench}, POPE \citep{li2023pope}, and MME \citep{fu2025mme}. The split and metric used for each benchmark are specified below.

\textbf{Benchmarks.}
Table~\ref{tab:app_benchmarks} lists the ten benchmarks used in our evaluation.
Every method, backbone, and budget configuration is evaluated on the identical
splits reported there, so that all comparisons in this paper are within-set
comparisons. All metrics are re-implemented following the official evaluation
protocol of each benchmark, and the same implementation is used for every
method, so that reported scores are directly comparable.

\begin{table}[!htbp]
\caption{Evaluation benchmarks. All methods share the same sets and the same metric implementation.}
    \label{tab:app_benchmarks}
    \centering
    \small
    \setlength{\tabcolsep}{5pt}
    \begin{tabular}{@{}lllll@{}}
        \toprule
        Group & Benchmark & \#Samples & Split & Metric\\
        \midrule
        \multirow{5}{*}{OCR / document / chart}
        & ChartQA      & 2{,}500   & test & relaxed accuracy\\
        & DocVQA       & 5{,}349   & validation & ANLS\\
        & TextVQA      & 5{,}000   & validation & VQA accuracy\\
        & HR-Bench 4K  & 800       & full & option accuracy\\
        & HR-Bench 8K  & 800       & full & option accuracy\\
        \midrule
        \multirow{5}{*}{General VQA / perception}
        & GQA          & 12{,}578  & testdev-balanced & exact-match accuracy\\
        & VQAv2        & 214{,}354 & validation & VQA accuracy\\
        & MMBench-EN   & 4{,}377   & dev & option accuracy\\
        & POPE         & 9{,}000   & adv./pop./rand. & F1\\
        & MME          & 2{,}374   & test & official MME score\\
        \bottomrule
    \end{tabular}
\end{table}

\textbf{Metrics.}
Each benchmark is scored with its official metric definition, summarized below
and listed per benchmark in Table~\ref{tab:app_benchmarks}. Throughout, $p$
denotes the model prediction after extraction and $g$ the reference answer.

\begin{itemize}
\item \textbf{Relaxed accuracy} (ChartQA~\citep{masry2022chartqa}). Introduced
for numeric chart question answering~\citep{methani2020plotqa}: a numeric
prediction is counted as correct if its relative error with respect to the
reference is at most $5\%$, while non-numeric predictions require exact string
match. The reported score is the fraction of correct predictions.

\item \textbf{Average Normalized Levenshtein Similarity (ANLS)}
(DocVQA~\citep{mathew2021docvqa}). Proposed for text-based visual question
answering~\citep{biten2019scenetext} to tolerate optical-character-recognition
errors while penalizing semantically wrong answers. With $d_{L}$ the
Levenshtein edit distance~\citep{levenshtein1966binary} and
$\mathrm{NL}(p,g)=d_{L}(p,g)/\max(|p|,|g|)$, the per-question score is
$1-\mathrm{NL}(p,g)$ if $\mathrm{NL}(p,g)<\tau$ and $0$ otherwise, with
$\tau=0.5$; the reported score averages over all questions. Multiple reference
answers are handled by taking the maximum over references.

\item \textbf{VQA accuracy} (VQAv2~\citep{goyal2017vqav2},
TextVQA~\citep{singh2019textvqa}). The soft agreement metric
of~\cite{antol2015vqa}: $\mathrm{Acc}(p)=\min\!\left(n(p)/3,\,1\right)$, where
$n(p)$ is the number of the ten human annotators whose answer equals $p$ after
the official normalization of articles, punctuation, and number words. The final
value averages this quantity over all ten nine-annotator subsets.

\item \textbf{Exact-match accuracy} (GQA~\citep{hudson2019gqa}). Answers are
drawn from a closed vocabulary, so a prediction is correct only if it matches
the reference exactly after the official normalization.

\item \textbf{Option accuracy} (MMBench-EN~\citep{liu2024mmbench}, HR-Bench 4K
and HR-Bench 8K~\citep{wang2025hrbench}). Both benchmarks are multiple choice
and are scored under the cyclic evaluation protocol
of~\citep{liu2024mmbench}: each question is presented once per cyclic
permutation of its options, and the question is counted as correct only if the
model selects the reference option under every permutation. This removes the
option-position bias that a single presentation would leave unmeasured.

\item \textbf{F1} (POPE~\citep{li2023pope}). Object-hallucination probing is cast
as binary yes/no classification with \texttt{yes} as the positive class; the
reported value is the harmonic mean of precision and recall, computed separately
for the random, popular, and adversarial sampling settings.

\item \textbf{Official MME score} (MME~\citep{fu2025mme}). Each of the fourteen
subtasks contributes $\mathrm{acc}+\mathrm{acc}^{+}$, where $\mathrm{acc}$ is
per-question accuracy and $\mathrm{acc}^{+}$ is the fraction of images for which
both associated questions are answered correctly, each expressed as a
percentage. The perception score sums the ten perception subtasks and the
cognition score sums the four cognition subtasks.
\end{itemize}

Because the evaluated models emit free-form dialogue rather than bare answers, a
shared extraction stage removes conversational prefixes and other non-answer
text before scoring, and every evaluation record stores the raw generation
alongside the extracted answer. This stage is necessary because the metrics
above operate at the string level. For $g=\texttt{1963}$ under ANLS, the raw
generation \texttt{The answer is 1963} yields $d_{L}=14$ and
$\max(|p|,|g|)=18$, hence $\mathrm{NL}=14/18>\tau$ and a score of $0$, whereas
the extracted answer \texttt{1963} yields $\mathrm{NL}=0$ and a score of $1$.
All metrics are re-implemented following the official protocol of each
benchmark, and the identical extraction stage and metric implementation are
applied to every compared method, so the benchmark metrics themselves are left
unchanged.

\textbf{Generation and image processing.}
Decoding is greedy for reproducibility, the generation length limit is set per benchmark, and the prompt suffix follows the convention of the training data of each backbone (a single-word/phrase instruction for open-ended VQA, an option-letter instruction for multiple choice) so that no arm loses points to instruction-format mismatch. Two backbone-side settings are fixed after measurement rather than by default: Qwen2.5-VL caps the visual sequence at $4096$ tokens, and InternVL3 enables dynamic tiling with the official maximum of $12$ patches. The latter is not the library default; leaving it off silently reduces a $4992\times7680$ image from $1792$ to $256$ visual tokens, which shows up only as a slightly lower score.

\textbf{Determinism and sharding.}
Evaluations are split into eight shards with deterministic construction and index-based Top-$k$ tie handling. All shards must load the same trained selector parameters, including the learned energy temperature $\tau_E$; construction seeds are not a substitute for checkpoint loading. First-minibatch temperature initialization belongs to training initialization and must not silently replace the learned value during evaluation.

\textbf{Correctness gates.}
The physical-removal path is checked at full retention against the unpruned generation path, with the retained count and position identifiers inspected. In particular, multidimensional positional information must be preserved for backbones that use it. G-Prune index selection is compared with its official implementation on the checked samples. These checks establish local implementation consistency rather than universal numerical equivalence. Run summaries surface per-sample failures so that a failed shard cannot silently reduce the set of evaluated samples.

\textbf{Efficiency measurement.}
Latency is measured separately from accuracy. The efficiency record uses $15$ samples each from GQA, ChartQA, DocVQA and HR-Bench 4K ($60$ total), with three warm-up runs and ten CUDA-event repetitions per sample. It reports prefill means, median selector overhead and greedy decoding over a forced $32$-token continuation. The reported comparison covers retention ratios $0.5$, $0.3$ and $0.1$, where the method's budget convention permits a run. Structural tokens remain outside the candidate budget; for example, $p=0.9$ can retain $11.6\%$ of all visual positions rather than exactly $10\%$.

Method-specific execution paths are important. One-shot selectors are timed after physical deletion or merging. For decoder-internal methods, the existing measurement uses one-token generation and a backbone-specific correction for the different language-model output-head computation. The recorded corrections are $+6.53$, $+4.70$, $+7.61$ and $-0.36$\,ms for LLaVA-NeXT, Qwen2.5-VL, InternVL3 and LLaVA-1.5, respectively; these are measurement-protocol corrections, not algorithmic speedups. Thus, the table compares the documented profiling paths, and raw first-token latency should not be confused with the corrected prefill quantity. Per-sample speedup ratios are aggregated, rather than calculated by dividing independently aggregated medians.

FLOPs are reported for the language model alone and for the end-to-end computation including visual processing. Decoder-internal pruning uses layer-specific sequence lengths in the FLOPs calculation. Neither FLOPs reduction nor prefill speedup is a measurement of complete-answer wall-clock latency. Selector overhead, sequence length and execution path all affect the observed timing.

\section{Baseline Implementations and Budget Accounting}
\label{app:baselines}

\textbf{Why every baseline is re-run.}
Every baseline is reproduced from the authors' open-source implementation and
published method specification inside our common evaluation pipeline, rather
than transcribed from the scores reported in the original papers. All methods
are thus measured under the same splits, the same metric implementation, and the
same answer-extraction stage. Re-running is necessary because published scores
depend on implementation details that the papers do not fully determine,
including the repository revision, the decoding configuration, the numerical
precision, and the answer post-processing; scores obtained along different
execution paths are therefore not directly comparable even when the benchmark
and the split coincide. Our baseline numbers may consequently deviate slightly
from the published ones, but the differences we report between methods are
attributable to the methods themselves rather than to their evaluation
harnesses.

\textbf{Budget accounting.}
One-shot methods reduce the visual sequence once before the language model by deletion or merging, so every layer processes the retained fraction $\rho$. FastV, PyramidDrop and SparseVLM prune inside the decoder, and their early layers still process the full sequence. Aligning budgets by the final token count would therefore credit them with compute they do not save, and the discrepancy grows with the pruning rate. We align by layer-averaged retention:
\begin{align}
    \text{FastV:}\quad & \bar\rho=\frac{K+(L-K)r}{L},
    &
    \text{PyramidDrop:}\quad & \bar\rho=\frac1S\sum_{j=0}^{S-1}\lambda^{j}.
    \label{eq:app_budget_accounting}
\end{align}
We solve for the method-specific parameter $r$ or $\lambda$ so that $\bar\rho$ equals the target budget. For the latter, we use bisection because the sum is monotone on $\lambda\in[0,1]$. SparseVLM is handled the same way, with recycled tokens counted as retained because they occupy sequence positions and are attended to. Under this convention, nominal $90\%$ pruning for FastV ($L=28$, $K=2$) corresponds to $16.4\%$ layer-averaged retention. Runs under the nominal convention are archived rather than deleted, so the question of whether the accounting changes any conclusion can be answered with data.

\textbf{Compared methods and evaluation coverage.}
Table~\ref{tab:app_baseline_deviations} summarizes the roles of the evaluated baselines. Table~\ref{tab:app_baseline_coverage} distinguishes unavailable evaluation coverage from the mathematical budget floor of staged pruning. Unreported entries do not imply that every missing backbone--method combination is architecturally impossible.

\begin{table}[!htbp]
\centering\small
\caption{Evaluated baseline roles. Implementations follow the corresponding open-source releases and method specifications; measurement and budget conventions are stated separately.}
\label{tab:app_baseline_deviations}
\begin{tabular}{@{}lp{0.69\textwidth}@{}}
\toprule
Method & Role in the comparison\\\midrule
G-Prune & Visual graph propagation and ranking.\\
FastV & Decoder-attention-based token selection.\\
PyramidDrop & Staged removal within the decoder.\\
SparseVLM & Text-informed visual token sparsification.\\
ToMe & Token merging rather than pure deletion.\\
VisionZip & Visual importance and contextual compression.\\
FasterVLM & Class-attention-based token deletion.\\
LLaVA-PruMerge & Selection combined with merging.\\
\bottomrule
\end{tabular}
\end{table}

\begin{table}[!htbp]
\centering\small
\caption{Reported evaluation coverage. Budget ranges use pruning rate $p$.}
\label{tab:app_baseline_coverage}
\begin{tabular}{@{}p{0.27\textwidth}p{0.14\textwidth}p{0.12\textwidth}p{0.37\textwidth}@{}}
\toprule
Method & Backbones & Budgets & Scope\\\midrule
FastV, ToMe, G-Prune & All four & $0.1$--$0.9$ & Evaluated across the reported grid.\\
PyramidDrop, SparseVLM & All four & $0.1$--$0.7$ & Under the implemented staged schedule, layer-averaged retention has a floor of $1/S=0.25$.\\
FasterVLM, VisionZip, PruMerge & LLaVA-1.5 & $0.1$--$0.9$ & Reported for this backbone; the present comparison does not evaluate their other backbone integrations.\\
\bottomrule
\end{tabular}
\end{table}

\section{Training Configuration}
\label{app:training_setup}

\textbf{What was trained.}
The selector is trained end-to-end through LLaVA-NeXT-8B. The vision encoder, projector and language model remain frozen; only the selector's approximately $4.5$M parameters are updated. Training uses only next-token cross-entropy at target-answer positions, as specified in Appendix~\ref{app:training}. Gradients from the task loss reach the selector through the frozen generation path and the multiplicative gate in Eq.~\eqref{eq:method_training_gate}.

Because the mask is computed before the language-model forward pass, the selector's textual context excludes the target answer. For single-turn samples, supervised positions are removed. For multi-turn samples, one target turn is sampled per epoch; loss is evaluated on that turn, and the context contains the system prompt, completed turns and current question. The information boundary is defined by token positions rather than answer-string matching.

\textbf{Two-stage retention-ratio training.}
Table~\ref{tab:app_training_stages} summarizes the training schedule. Stage A uses the strong-signal subset for approximately $0.3$ of an epoch at fixed $\rho=0.9$ and STE temperature $\beta=2.0$. This stage provides initial training under a loose token budget. Stage B uses the source-weighted $779$K mixture for approximately $2$ epochs. Each cycle consists of four steps in which $\rho$ is sampled independently and uniformly from $\{0.5,0.3,0.1\}$, followed by one additional step at $\rho=0.9$. Thus, the loose-budget step occurs once per five-step cycle and does not replace one of the four randomly sampled steps. The STE temperature follows cosine annealing from $2.0$ to $0.1$ during Stage B.

The sampled retention ratio enters the selector through $g(\rho)$. The training ratios are $\{0.9,0.5,0.3,0.1\}$, whereas evaluation covers seven retention ratios, $\{0.9,0.7,0.5,0.4,0.3,0.2,0.1\}$. All seven use the same trained checkpoint, including the ratios not sampled during training.

\begin{table}[!htbp]
\centering
\caption{Two-stage selector training. Retention is denoted by $\rho$; pruning rate is $p=1-\rho$. Approximate epoch counts follow the training configuration.}
\label{tab:app_training_stages}
\small
\begin{tabular}{@{}>{\raggedright\arraybackslash}p{0.17\textwidth}>{\raggedright\arraybackslash}p{0.29\textwidth}>{\raggedright\arraybackslash}p{0.44\textwidth}@{}}
\toprule
Setting & Stage A: initial alignment & Stage B: ratio conditioning \\
\midrule
Data & Strong-signal subset & Source-weighted $779$K mixture \\
Duration & Approximately $0.3$ of an epoch & Approximately $2$ epochs \\
Retention & Fixed $\rho=0.9$ & Four uniformly sampled steps from $\{0.5,0.3,0.1\}$, then one additional step at $\rho=0.9$ \\
STE temperature & Fixed $\beta=2.0$ & Cosine annealing, $2.0\rightarrow0.1$ \\
Loss & Target-answer cross-entropy & Target-answer cross-entropy \\
\bottomrule
\end{tabular}
\end{table}

\textbf{Source-weighted data sampling.}
The instruction mixture assigns different sampling weights to task categories according to their visual-selection signal (Table~\ref{tab:app_source_sampling}). Fine-grained localization, OCR, chart understanding and grounding/referring tasks receive weight $2.5$; general visual question answering and GPT-4V-generated visual question-answer or description data receive weight $1.0$; generic descriptions, summaries and conversations in which the image is peripheral receive weight $0.5$. Text-only samples without an image field are filtered by the data loader, and the selector retains a short-circuit path for zero visual tokens. These are sampling weights, not fixed proportions of the resulting training stream.

\begin{table}[!htbp]
\centering
\caption{Relative sampling weights by task category. Descriptions identify task types rather than an inventory of dataset names or splits.}
\label{tab:app_source_sampling}
\small
\begin{tabular}{@{}>{\raggedright\arraybackslash}p{0.15\textwidth}>{\raggedright\arraybackslash}p{0.65\textwidth}r@{}}
\toprule
Category & Task characteristics & Weight \\
\midrule
Strong & Fine-grained localization; OCR and text recognition; chart understanding; grounding and referring tasks & $2.5$ \\
Regular & General VQA; GPT-4V-generated visual questions, answers and descriptions & $1.0$ \\
Weak & Generic descriptions and summaries; dialogue with peripheral visual content & $0.5$ \\
Excluded & Text-only samples without an image field & $0$ \\
\bottomrule
\end{tabular}
\end{table}

\textbf{Optimization.}
Optimization uses AdamW with learning rate $2\times10^{-4}$, weight decay $0.01$, $1000$ learning-rate warm-up steps, cosine decay and gradient clipping at $1.0$. Positional embeddings and learnable scalars use a $0.1\times$ learning rate. Training uses an effective batch size of $64$--$128$, bfloat16, eager attention and DDP on two $96$\,GB GPUs. The learning-rate warm-up and Stage A specify different aspects of the training schedule.


\textbf{How the reported entries were produced.}
All main comparison figures and tables use this trained checkpoint, evaluated on the three remaining backbones without retraining. Reproducing the displayed SheafPrune envelope requires evaluating that checkpoint over $\delta\in\{0,0.5,1,2\}$ and reproducing the original per-point selection rule. The aggregate tables alone do not establish that this choice used a held-out validation set, so no held-out selection or tuning-free deployment claim is made. A fixed deployment curve would instead use one precommitted $\delta$ across the complete grid. The fixed-configuration component study uses default $\delta=1$ except in the no-correction arm.

\textbf{Scope of the training evidence.}
The method is trained with target-answer cross-entropy through the frozen backbone. The uniform-score ablation changes the initialization supplied to the graph operator; it is not a comparison between trained and untrained checkpoints. The current component results therefore do not independently establish the necessity of every trainable submodule, nor do they establish that training is unnecessary.

\section{Complete Results}
\label{app:rawscores}

Table~\ref{tab:exp_all} contains the complete reported main-comparison grid: for each backbone it lists the evaluated methods and pruning rates, original scores on the ten benchmarks, the ten-benchmark mean and macro-average retention, and --- at the three profiled budgets --- prefill latency, latency reduction, decoding throughput and end-to-end FLOPs reduction. Table~\ref{tab:exp_main} of Section~\ref{sec:exp_main} is its $p=0.7$ slice. Retention is computed per benchmark and then averaged across benchmarks, so each benchmark contributes on the same footing despite different native score scales; raw MME scores remain on their native scale in the MME column. Empty entries reflect either the implemented staged schedule's budget floor or combinations not evaluated here, as detailed in Appendix~\ref{app:baselines}. Missing values are not treated as zero. SheafPrune entries follow the per-point best-$\delta$ reporting protocol in Appendix~\ref{app:training_setup}.

\begingroup
\setlength{\tabcolsep}{1.0pt}
\renewcommand{\arraystretch}{0.80}
\scriptsize
\setlength{\LTcapwidth}{\textwidth}
\begin{longtable}{@{}lc|cc|ccc|ccc|cc|cc|ccccc@{}}
    \caption{\textbf{Complete results.} Every backbone, method and pruning rate in one table. Benchmark columns are the original scores of the final result workbook: Chart, Doc, Text, MMB and VQA2 are ChartQA (relaxed accuracy), DocVQA (ANLS), TextVQA, MMBench-EN and VQAv2; POPE is F1, MME keeps its native scale, the remaining benchmarks use accuracy-based metrics, and PruMerge is LLaVA-PruMerge. ``Avg'' is the ten-benchmark mean with MME rescaled as score$/28$; ``RA'' is the ten-benchmark macro-average retention (\%) relative to the unpruned model of the same backbone, i.e.\ each benchmark is normalized first and the ratios are then averaged, which is the quantity every claim in the paper is stated in. ``lat'' is the measured prefill latency (ms) of the complete method --- the forward pass plus that method's own selection cost --- and ``LR'' the latency reduction (\%) it gives against the unpruned prefill of the same backbone; ``tok/s'' is greedy decoding throughput and ``FLOPs'' the end-to-end FLOPs reduction (\%) including the frozen vision tower. Efficiency was profiled at $p\in\{0.5,0.7,0.9\}$ on one $60$-sample batch shared by all four backbones, so it is a per-backbone quantity and ``--'' in those columns marks a budget that was not profiled. SparseVLM and PyramidDrop prune inside the decoder in stages and their layer-averaged retention cannot fall below $1/S=25\%$, so they have no $p\in\{0.8,0.9\}$ entry at all; FasterVLM, VisionZip and LLaVA-PruMerge are reported only on LLaVA-1.5 in this evaluation; other backbone integrations are not evaluated here. Best pruning method per backbone, pruning rate and column in bold.}
    \label{tab:exp_all}\\
    \toprule
    Method & $p$ & \multicolumn{2}{c|}{General VQA} & \multicolumn{3}{c|}{MLLM benchmarks} & \multicolumn{3}{c|}{Text-oriented VQA} & \multicolumn{2}{c|}{High-resolution} & \multicolumn{2}{c|}{Aggregate} & \multicolumn{5}{c}{Efficiency}\\
    \cmidrule(lr){3-4}\cmidrule(lr){5-7}\cmidrule(lr){8-10}\cmidrule(lr){11-12}\cmidrule(lr){13-14}\cmidrule(lr){15-19}
    & (\%) & GQA & VQA2 & MME & POPE & MMB & Chart & Doc & Text & HR-4K & HR-8K & Avg & RA & lat & $\times$ & LR & tok/s & FLOPs\\
    \midrule
    \endfirsthead
    \multicolumn{19}{@{}l}{\scriptsize\emph{Table~\ref{tab:exp_all} (continued).}}\\
    \toprule
    Method & $p$ & \multicolumn{2}{c|}{General VQA} & \multicolumn{3}{c|}{MLLM benchmarks} & \multicolumn{3}{c|}{Text-oriented VQA} & \multicolumn{2}{c|}{High-resolution} & \multicolumn{2}{c|}{Aggregate} & \multicolumn{5}{c}{Efficiency}\\
    \cmidrule(lr){3-4}\cmidrule(lr){5-7}\cmidrule(lr){8-10}\cmidrule(lr){11-12}\cmidrule(lr){13-14}\cmidrule(lr){15-19}
    & (\%) & GQA & VQA2 & MME & POPE & MMB & Chart & Doc & Text & HR-4K & HR-8K & Avg & RA & lat & $\times$ & LR & tok/s & FLOPs\\
    \midrule
    \endhead
    \bottomrule
    \endlastfoot
    \multicolumn{19}{@{}l}{\textbf{LLaVA-NeXT-8B}}\\
    \midrule
    No prune & $0$ & 65.0 & 82.1 & 1074.21 & 83.6 & 77.0 & 69.7 & 72.6 & 66.7 & 51.2 & 46.1 & 65.2 & -- & 149.4 & -- & -- & 66.5 & 0.0\\
    \addlinespace[1.2pt]
    Random & $10$ & 64.9 & 82.4 & 1076.44 & 83.0 & \textbf{77.2} & 67.8 & 70.8 & 66.3 & 51.5 & 46.4 & 64.9 & 99.5 & -- & -- & -- & -- & --\\
    Random & $30$ & 64.5 & 82.1 & 1056.92 & 82.9 & 76.8 & 62.4 & 62.1 & 64.0 & 50.5 & 45.2 & 62.8 & 96.4 & -- & -- & -- & -- & --\\
    Random & $50$ & 64.4 & 80.0 & 1075.43 & 82.2 & \textbf{77.3} & 56.5 & 51.3 & 57.0 & 47.1 & 44.6 & 59.9 & 92.1 & -- & -- & -- & -- & --\\
    Random & $60$ & 63.9 & 80.0 & 1070.32 & 81.7 & 76.9 & 49.6 & 46.6 & 54.5 & 48.6 & 42.5 & 58.2 & 89.7 & -- & -- & -- & -- & --\\
    Random & $70$ & 63.8 & 79.3 & 1050.99 & 80.7 & \textbf{77.2} & 43.6 & 39.9 & 49.9 & 46.0 & 42.2 & 56.0 & 86.3 & -- & -- & -- & -- & --\\
    Random & $80$ & 63.1 & 77.7 & 1045.53 & 78.3 & 74.6 & 37.2 & 31.1 & 42.5 & 45.8 & 40.1 & 52.8 & 81.6 & -- & -- & -- & -- & --\\
    Random & $90$ & 59.1 & 72.5 & 1034.32 & 73.9 & 73.0 & 27.9 & 22.2 & 31.4 & 40.6 & 36.0 & 47.4 & 73.4 & -- & -- & -- & -- & --\\
    \addlinespace[1.2pt]
    ToMe & $10$ & 65.0 & 82.6 & 1073.51 & 83.6 & 76.9 & 69.6 & 72.4 & 66.4 & \textbf{51.9} & 46.5 & \textbf{65.3} & \textbf{100.2} & -- & -- & -- & -- & --\\
    ToMe & $30$ & 65.0 & 82.1 & 1055.83 & 83.5 & 76.4 & 69.1 & 69.8 & 66.0 & 51.2 & 45.8 & 64.6 & 99.1 & -- & -- & -- & -- & --\\
    ToMe & $50$ & \textbf{65.7} & 81.6 & 1064.97 & 83.4 & 76.9 & 57.4 & 57.1 & 63.0 & 50.0 & 42.1 & 61.5 & 94.4 & \textbf{78.6} & \textbf{1.90} & \textbf{47.4} & 66.2 & 48.1\\
    ToMe & $60$ & \textbf{65.4} & 81.8 & 1053.85 & 82.3 & 77.1 & 57.0 & 56.6 & 62.5 & 50.6 & 42.9 & 61.4 & 94.2 & -- & -- & -- & -- & --\\
    ToMe & $70$ & 64.5 & 80.8 & 1057.58 & 82.4 & 76.3 & 52.1 & 48.7 & 60.5 & 50.0 & 42.5 & 59.6 & 91.6 & \textbf{52.6} & \textbf{2.84} & \textbf{64.8} & 68.1 & 66.3\\
    ToMe & $80$ & 63.9 & 79.3 & 1057.82 & 80.3 & 75.4 & 38.8 & 39.3 & 51.5 & 49.9 & 41.6 & 55.8 & 86.2 & -- & -- & -- & -- & --\\
    ToMe & $90$ & 62.5 & 75.9 & \textbf{1043.83} & 77.7 & 73.0 & 27.5 & 28.6 & 38.8 & 44.1 & 39.4 & 50.5 & 78.2 & 32.6 & 4.58 & 78.2 & \textbf{68.8} & 84.0\\
    \addlinespace[1.2pt]
    FastV & $10$ & \textbf{65.2} & \textbf{82.6} & 1057.88 & 83.6 & 76.9 & 69.6 & 72.6 & 66.5 & 51.6 & 46.4 & 65.3 & 100.0 & -- & -- & -- & -- & --\\
    FastV & $30$ & 64.7 & \textbf{82.6} & 1061.31 & 83.2 & 77.0 & 69.0 & 70.1 & 66.4 & 50.8 & 46.5 & 64.8 & 99.3 & -- & -- & -- & -- & --\\
    FastV & $50$ & 64.5 & 82.3 & 1054.56 & 83.4 & 77.0 & 67.1 & 66.7 & 65.2 & 51.5 & 47.0 & 64.2 & 98.6 & 80.8 & 1.85 & 45.9 & 52.0 & 49.0\\
    FastV & $60$ & 64.0 & 82.1 & 1058.83 & 83.2 & 76.5 & 63.9 & 63.5 & 64.5 & 50.5 & 46.6 & 63.2 & 97.1 & -- & -- & -- & -- & --\\
    FastV & $70$ & 63.5 & 81.0 & 1056.34 & 82.5 & 76.1 & 59.2 & 55.2 & 61.7 & 51.4 & 45.5 & 61.4 & 94.5 & 55.7 & 2.68 & 62.7 & 51.7 & 67.5\\
    FastV & $80$ & 61.7 & 78.1 & 1036.65 & 80.8 & 75.1 & 48.8 & 42.2 & 56.5 & 48.6 & 44.0 & 57.3 & 88.4 & -- & -- & -- & -- & --\\
    FastV & $90$ & 53.4 & 70.0 & 953.06 & 65.5 & 69.7 & 23.9 & 18.4 & 38.9 & 41.2 & 39.6 & 45.5 & 70.9 & 33.0 & 4.53 & 77.9 & 51.4 & \textbf{85.4}\\
    \addlinespace[1.2pt]
    SparseVLM & $10$ & 64.7 & 82.4 & 1063.02 & \textbf{83.8} & 77.2 & 69.5 & 72.2 & 66.4 & 51.6 & \textbf{46.6} & 65.2 & 100.0 & -- & -- & -- & -- & --\\
    SparseVLM & $30$ & 64.1 & 82.3 & 1059.00 & 83.8 & 76.7 & 69.4 & 72.5 & 66.3 & \textbf{51.6} & 46.6 & 65.1 & 99.8 & -- & -- & -- & -- & --\\
    SparseVLM & $50$ & 64.4 & 82.0 & 1058.22 & \textbf{84.5} & 76.6 & 68.1 & 71.0 & 64.1 & \textbf{52.0} & 46.9 & 64.7 & 99.3 & 84.4 & 1.77 & 43.5 & 51.6 & 49.7\\
    SparseVLM & $60$ & 62.8 & 81.1 & 1055.18 & \textbf{84.2} & 77.0 & 64.0 & 65.4 & 59.8 & \textbf{52.5} & \textbf{47.4} & 63.2 & 97.1 & -- & -- & -- & -- & --\\
    SparseVLM & $70$ & 58.5 & 76.5 & 1039.67 & 83.0 & 76.8 & 45.0 & 46.4 & 46.9 & 51.0 & 46.1 & 56.7 & 87.8 & 63.3 & 2.36 & 57.6 & 51.2 & 67.9\\
    SparseVLM & $80$ & -- & -- & -- & -- & -- & -- & -- & -- & -- & -- & -- & -- & -- & -- & -- & -- & --\\
    SparseVLM & $90$ & -- & -- & -- & -- & -- & -- & -- & -- & -- & -- & -- & -- & -- & -- & -- & -- & --\\
    \addlinespace[1.2pt]
    PyramidDrop & $10$ & 65.1 & 82.3 & 1062.25 & 83.6 & 77.0 & 69.7 & 71.9 & 66.4 & 51.4 & 46.5 & 65.2 & 99.9 & -- & -- & -- & -- & --\\
    PyramidDrop & $30$ & \textbf{65.1} & 82.2 & 1060.00 & \textbf{83.8} & \textbf{77.2} & 69.5 & 72.2 & 66.4 & 51.2 & \textbf{46.8} & \textbf{65.2} & 100.0 & -- & -- & -- & -- & --\\
    PyramidDrop & $50$ & 64.3 & \textbf{82.4} & 1057.51 & 83.8 & 76.9 & \textbf{69.6} & 71.3 & 66.4 & 51.6 & \textbf{47.0} & 65.1 & 99.8 & 84.9 & 1.76 & 43.2 & 50.9 & \textbf{50.2}\\
    PyramidDrop & $60$ & 64.2 & \textbf{82.3} & 1054.05 & 83.8 & \textbf{77.2} & 67.4 & 68.2 & 64.8 & 52.4 & 46.8 & 64.5 & 98.9 & -- & -- & -- & -- & --\\
    PyramidDrop & $70$ & 62.2 & 80.8 & 1057.45 & \textbf{85.0} & 76.1 & 45.4 & 49.3 & 55.5 & \textbf{52.1} & 45.6 & 59.0 & 91.0 & 63.0 & 2.37 & 57.8 & 50.9 & \textbf{68.0}\\
    PyramidDrop & $80$ & -- & -- & -- & -- & -- & -- & -- & -- & -- & -- & -- & -- & -- & -- & -- & -- & --\\
    PyramidDrop & $90$ & -- & -- & -- & -- & -- & -- & -- & -- & -- & -- & -- & -- & -- & -- & -- & -- & --\\
    \addlinespace[1.2pt]
    G-Prune & $10$ & 64.7 & 82.0 & \textbf{1085.27} & 83.4 & 76.9 & 69.6 & 72.5 & \textbf{67.1} & 51.2 & 45.9 & 65.2 & 100.0 & -- & -- & -- & -- & --\\
    G-Prune & $30$ & 64.5 & 82.1 & 1071.84 & 83.3 & 77.0 & \textbf{69.7} & 71.7 & \textbf{67.2} & 50.1 & 46.2 & 65.0 & 99.6 & -- & -- & -- & -- & --\\
    G-Prune & $50$ & 64.9 & 82.2 & \textbf{1083.51} & 83.2 & 76.3 & 68.9 & 70.4 & 66.2 & 50.1 & 46.5 & 64.7 & 99.3 & 79.5 & 1.88 & 46.8 & 66.4 & 48.1\\
    G-Prune & $60$ & 65.1 & 82.0 & \textbf{1077.27} & 83.8 & 75.8 & 68.0 & 68.6 & 66.1 & 51.4 & 46.2 & 64.6 & 99.1 & -- & -- & -- & -- & --\\
    G-Prune & $70$ & \textbf{64.9} & 81.6 & 1059.47 & 83.6 & 75.7 & 66.5 & 64.8 & 65.6 & 51.2 & 45.8 & 63.8 & 97.9 & 52.8 & 2.83 & 64.7 & 67.0 & 66.3\\
    G-Prune & $80$ & 63.8 & \textbf{80.9} & 1055.26 & 83.5 & 75.6 & 63.4 & 59.0 & 62.7 & 50.6 & 45.6 & 62.3 & 95.7 & -- & -- & -- & -- & --\\
    G-Prune & $90$ & 62.1 & 76.7 & 1015.24 & 79.8 & 73.9 & 52.7 & 43.0 & 59.1 & 48.5 & 43.9 & 57.6 & 88.8 & 32.3 & 4.62 & 78.4 & 67.0 & 84.0\\
    \addlinespace[1.2pt]
    \textbf{SheafPrune} & $10$ & 65.0 & 82.1 & 1074.21 & 83.6 & 77.0 & \textbf{69.7} & \textbf{72.6} & 66.7 & 51.2 & 46.1 & 65.2 & 100.0 & -- & -- & -- & -- & --\\
    \textbf{SheafPrune} & $30$ & 65.0 & 82.1 & \textbf{1074.19} & 83.6 & 77.0 & 69.7 & \textbf{72.6} & 66.7 & 51.2 & 46.1 & 65.2 & \textbf{100.0} & -- & -- & -- & -- & --\\
    \textbf{SheafPrune} & $50$ & 65.0 & 82.1 & 1074.21 & 83.6 & 76.9 & 69.5 & \textbf{72.2} & \textbf{66.7} & 51.2 & 46.1 & \textbf{65.2} & \textbf{99.9} & 81.6 & 1.83 & 45.4 & \textbf{67.9} & 48.1\\
    \textbf{SheafPrune} & $60$ & 65.0 & 82.1 & 1074.21 & 83.6 & 76.6 & \textbf{69.0} & \textbf{70.8} & \textbf{66.6} & 51.2 & 46.0 & \textbf{64.9} & \textbf{99.5} & -- & -- & -- & -- & --\\
    \textbf{SheafPrune} & $70$ & 64.8 & \textbf{82.1} & \textbf{1074.21} & 83.6 & 76.2 & \textbf{67.5} & \textbf{67.2} & \textbf{66.1} & 51.2 & \textbf{46.1} & \textbf{64.3} & \textbf{98.7} & 55.3 & 2.70 & 63.0 & \textbf{68.3} & 66.4\\
    \textbf{SheafPrune} & $80$ & \textbf{64.1} & 80.8 & \textbf{1065.10} & \textbf{83.6} & \textbf{76.3} & \textbf{64.6} & \textbf{62.2} & \textbf{64.3} & \textbf{51.0} & \textbf{46.1} & \textbf{63.1} & \textbf{97.0} & -- & -- & -- & -- & --\\
    \textbf{SheafPrune} & $90$ & \textbf{62.7} & \textbf{78.0} & 1022.89 & \textbf{80.8} & \textbf{74.2} & \textbf{54.1} & \textbf{46.6} & \textbf{60.0} & \textbf{49.7} & \textbf{44.1} & \textbf{58.7} & \textbf{90.4} & \textbf{29.0} & \textbf{5.15} & \textbf{80.6} & 67.3 & 84.0\\
    \midrule
    \multicolumn{19}{@{}l}{\textbf{Qwen2.5-VL-7B}}\\
    \midrule
    No prune & $0$ & 58.7 & 83.1 & 1242.65 & 84.3 & 83.3 & 87.8 & 94.2 & 84.3 & 71.4 & 63.9 & 75.5 & -- & 113.6 & -- & -- & 71.0 & 0.0\\
    \addlinespace[1.2pt]
    Random & $10$ & 58.5 & 83.0 & 1238.03 & \textbf{84.5} & 83.1 & 86.8 & 93.9 & 84.2 & 70.9 & 63.6 & 75.3 & 99.7 & -- & -- & -- & -- & --\\
    Random & $30$ & 58.3 & 82.7 & 1231.26 & 83.7 & 83.5 & 84.1 & 93.0 & 81.9 & 69.6 & 62.5 & 74.3 & 98.5 & -- & -- & -- & -- & --\\
    Random & $50$ & 57.9 & 81.5 & 1229.30 & 82.9 & 81.8 & 77.8 & 88.3 & 78.1 & 69.8 & 60.2 & 72.2 & 95.9 & -- & -- & -- & -- & --\\
    Random & $60$ & 57.7 & 80.6 & 1205.57 & 83.0 & \textbf{83.2} & 71.3 & 85.4 & 75.2 & 66.5 & 59.4 & 70.5 & 93.8 & -- & -- & -- & -- & --\\
    Random & $70$ & 56.7 & 78.1 & 1178.81 & 82.3 & \textbf{82.3} & 64.7 & 77.3 & 71.0 & 65.2 & 57.4 & 67.7 & 90.3 & -- & -- & -- & -- & --\\
    Random & $80$ & 55.7 & 76.2 & 1169.27 & 80.3 & \textbf{80.3} & 52.6 & 63.6 & 64.6 & 63.4 & 53.6 & 63.2 & 84.9 & -- & -- & -- & -- & --\\
    Random & $90$ & 51.9 & 71.1 & 1092.10 & 74.1 & 77.9 & 37.7 & 44.5 & 49.2 & 58.6 & 51.9 & 55.6 & 75.5 & -- & -- & -- & -- & --\\
    \addlinespace[1.2pt]
    ToMe & $10$ & 58.7 & 83.3 & 1241.65 & 84.3 & 83.4 & 87.6 & 94.1 & 84.2 & 71.6 & 64.1 & 75.6 & 100.1 & -- & -- & -- & -- & --\\
    ToMe & $30$ & 58.4 & \textbf{83.2} & 1238.00 & 83.9 & 83.3 & 86.4 & 93.6 & 83.5 & 70.9 & 64.2 & 75.2 & 99.6 & -- & -- & -- & -- & --\\
    ToMe & $50$ & \textbf{58.8} & \textbf{82.1} & \textbf{1236.98} & 83.0 & 82.7 & 82.9 & 93.2 & 82.4 & 69.1 & 62.6 & 74.1 & 98.2 & \textbf{62.1} & \textbf{1.83} & \textbf{45.4} & 67.0 & 35.8\\
    ToMe & $60$ & \textbf{58.8} & \textbf{81.5} & \textbf{1224.74} & 83.9 & 82.4 & 79.3 & 90.7 & 81.3 & 67.9 & 63.6 & 73.3 & 97.3 & -- & -- & -- & -- & --\\
    ToMe & $70$ & \textbf{58.2} & \textbf{80.8} & \textbf{1214.42} & 83.3 & 81.3 & 73.1 & 84.7 & 79.5 & 68.1 & 60.9 & 71.3 & 94.9 & \textbf{43.4} & \textbf{2.62} & \textbf{61.8} & 68.1 & 49.5\\
    ToMe & $80$ & \textbf{56.7} & \textbf{78.7} & \textbf{1185.98} & 82.4 & 80.1 & 62.4 & 75.0 & 74.1 & 67.2 & 59.6 & 67.9 & 90.7 & -- & -- & -- & -- & --\\
    ToMe & $90$ & \textbf{54.7} & \textbf{74.8} & \textbf{1131.61} & 79.6 & \textbf{78.9} & 48.1 & 51.3 & \textbf{61.9} & 65.9 & 57.8 & 61.3 & 82.9 & 35.8 & 3.17 & 68.5 & 69.9 & 63.0\\
    \addlinespace[1.2pt]
    FastV & $10$ & 58.5 & 83.2 & 1233.22 & 84.1 & 83.2 & 87.6 & \textbf{94.4} & 84.1 & 71.6 & \textbf{64.4} & 75.5 & 100.0 & -- & -- & -- & -- & --\\
    FastV & $30$ & 57.9 & 82.7 & 1241.82 & 84.5 & 82.5 & 86.1 & 92.9 & 84.0 & \textbf{72.4} & \textbf{65.4} & 75.3 & 99.7 & -- & -- & -- & -- & --\\
    FastV & $50$ & 56.9 & 81.5 & 1236.19 & 84.2 & 82.5 & 80.6 & 90.1 & 83.2 & 72.2 & 63.5 & 73.9 & 98.0 & 64.5 & 1.76 & 43.2 & 46.4 & 35.9\\
    FastV & $60$ & 54.7 & 79.5 & 1218.42 & 82.0 & 81.9 & 74.5 & 84.4 & 82.9 & 70.9 & 63.8 & 71.8 & 95.4 & -- & -- & -- & -- & --\\
    FastV & $70$ & 51.7 & 76.8 & 1170.01 & 79.9 & 79.5 & 63.0 & 74.5 & \textbf{82.3} & 68.1 & 61.9 & 68.0 & 90.6 & 47.9 & 2.37 & 57.8 & 46.4 & 49.7\\
    FastV & $80$ & 47.7 & 71.5 & 1106.28 & 74.9 & 74.5 & 40.4 & 55.2 & \textbf{76.6} & 63.5 & 57.9 & 60.2 & 81.0 & -- & -- & -- & -- & --\\
    FastV & $90$ & 38.0 & 44.8 & 869.22 & 18.1 & 56.9 & 15.9 & 25.4 & 48.1 & 53.6 & 45.4 & 37.7 & 52.7 & 30.4 & 3.74 & 73.3 & 46.1 & \textbf{63.1}\\
    \addlinespace[1.2pt]
    SparseVLM & $10$ & 58.7 & 83.1 & 1244.32 & 84.5 & 83.0 & 87.3 & 94.0 & 84.2 & \textbf{72.0} & 64.0 & 75.5 & 100.0 & -- & -- & -- & -- & --\\
    SparseVLM & $30$ & 58.1 & 82.4 & 1241.01 & 84.9 & \textbf{83.5} & 85.0 & 93.9 & 83.6 & 71.9 & 64.2 & 75.2 & 99.6 & -- & -- & -- & -- & --\\
    SparseVLM & $50$ & 56.5 & 80.3 & 1229.35 & 83.9 & \textbf{82.9} & 77.7 & 92.3 & 83.7 & 71.9 & 64.5 & 73.8 & 97.8 & 72.8 & 1.56 & 35.9 & 45.9 & 36.7\\
    SparseVLM & $60$ & 54.7 & 76.5 & 1221.25 & 83.1 & 80.6 & 65.5 & 87.3 & 82.0 & \textbf{72.8} & 64.2 & 71.0 & 94.6 & -- & -- & -- & -- & --\\
    SparseVLM & $70$ & 49.2 & 68.9 & 1124.35 & 76.2 & 76.9 & 27.3 & 67.3 & 66.7 & 69.5 & 63.0 & 60.5 & 81.8 & 57.1 & 1.99 & 49.7 & 46.0 & 50.3\\
    SparseVLM & $80$ & -- & -- & -- & -- & -- & -- & -- & -- & -- & -- & -- & -- & -- & -- & -- & -- & --\\
    SparseVLM & $90$ & -- & -- & -- & -- & -- & -- & -- & -- & -- & -- & -- & -- & -- & -- & -- & -- & --\\
    \addlinespace[1.2pt]
    PyramidDrop & $10$ & 58.5 & 83.3 & 1241.84 & 84.1 & 82.8 & 87.5 & 94.0 & 84.2 & 71.4 & 64.2 & 75.4 & 99.9 & -- & -- & -- & -- & --\\
    PyramidDrop & $30$ & 58.4 & 82.8 & \textbf{1243.73} & 85.0 & 83.0 & 86.7 & 94.2 & 83.7 & 72.0 & 64.4 & 75.5 & 100.0 & -- & -- & -- & -- & --\\
    PyramidDrop & $50$ & 56.9 & 81.9 & 1215.19 & \textbf{84.6} & 81.0 & 81.7 & 92.3 & \textbf{84.4} & 71.6 & 64.8 & 74.3 & 98.4 & 71.9 & 1.58 & 36.7 & 45.6 & \textbf{37.1}\\
    PyramidDrop & $60$ & 54.5 & 79.8 & 1166.81 & 82.2 & 78.3 & 73.9 & 88.9 & \textbf{83.5} & 70.5 & 64.0 & 71.7 & 95.1 & -- & -- & -- & -- & --\\
    PyramidDrop & $70$ & 48.1 & 71.5 & 1018.31 & 74.9 & 71.3 & 41.0 & 67.6 & 69.1 & 63.2 & 54.6 & 59.8 & 79.9 & 56.8 & 2.00 & 50.0 & 45.7 & \textbf{50.4}\\
    PyramidDrop & $80$ & -- & -- & -- & -- & -- & -- & -- & -- & -- & -- & -- & -- & -- & -- & -- & -- & --\\
    PyramidDrop & $90$ & -- & -- & -- & -- & -- & -- & -- & -- & -- & -- & -- & -- & -- & -- & -- & -- & --\\
    \addlinespace[1.2pt]
    G-Prune & $10$ & \textbf{59.1} & \textbf{83.3} & \textbf{1245.59} & 84.3 & \textbf{83.4} & 87.7 & 94.1 & \textbf{84.3} & 71.5 & 63.9 & \textbf{75.6} & \textbf{100.1} & -- & -- & -- & -- & --\\
    G-Prune & $30$ & 57.9 & 82.3 & 1242.56 & \textbf{85.0} & 83.3 & 86.4 & \textbf{94.3} & 84.2 & 72.0 & 63.9 & 75.4 & 99.8 & -- & -- & -- & -- & --\\
    G-Prune & $50$ & 55.7 & 80.2 & 1213.94 & 84.5 & 81.8 & 82.5 & 93.4 & 84.2 & \textbf{72.8} & \textbf{65.2} & 74.3 & 98.4 & 63.5 & 1.79 & 44.1 & 69.2 & 35.7\\
    G-Prune & $60$ & 53.9 & 78.5 & 1184.42 & 83.5 & 80.6 & 77.0 & 91.6 & 81.7 & 72.6 & \textbf{64.4} & 72.6 & 96.2 & -- & -- & -- & -- & --\\
    G-Prune & $70$ & 51.2 & 74.8 & 1115.52 & 80.1 & 78.5 & 66.8 & 88.0 & 78.4 & \textbf{72.8} & \textbf{64.5} & 69.5 & 92.2 & 44.0 & 2.58 & 61.2 & 67.5 & 49.5\\
    G-Prune & $80$ & 48.9 & 68.8 & 1011.88 & 73.5 & 76.0 & 51.3 & 76.2 & 71.4 & \textbf{70.8} & \textbf{63.5} & 63.7 & 84.8 & -- & -- & -- & -- & --\\
    G-Prune & $90$ & 43.5 & 60.1 & 928.41 & 61.8 & 67.2 & 28.7 & 52.6 & 52.0 & 68.0 & 60.0 & 52.7 & 71.4 & 30.1 & 3.78 & 73.5 & 67.2 & 62.9\\
    \addlinespace[1.2pt]
    \textbf{SheafPrune} & $10$ & 58.6 & 83.1 & 1242.65 & 84.2 & 83.3 & \textbf{87.8} & 94.2 & 84.3 & 71.4 & 63.9 & 75.5 & 100.0 & -- & -- & -- & -- & --\\
    \textbf{SheafPrune} & $30$ & \textbf{58.7} & 83.1 & 1242.52 & 84.3 & 83.2 & \textbf{87.7} & 94.2 & \textbf{84.3} & 71.4 & 63.9 & \textbf{75.5} & \textbf{100.0} & -- & -- & -- & -- & --\\
    \textbf{SheafPrune} & $50$ & 58.7 & 81.9 & 1223.50 & 84.3 & 82.3 & \textbf{84.2} & \textbf{93.4} & 83.4 & 71.4 & 63.8 & \textbf{74.7} & \textbf{99.0} & 65.7 & 1.73 & 42.2 & \textbf{72.0} & 35.8\\
    \textbf{SheafPrune} & $60$ & 58.1 & 80.8 & 1204.79 & \textbf{84.3} & 82.6 & \textbf{81.9} & \textbf{91.7} & 82.0 & 71.4 & 63.8 & \textbf{74.0} & \textbf{98.0} & -- & -- & -- & -- & --\\
    \textbf{SheafPrune} & $70$ & 57.0 & 80.0 & 1156.14 & \textbf{84.2} & 81.2 & \textbf{77.1} & \textbf{88.1} & 79.1 & 71.4 & 63.5 & \textbf{72.3} & \textbf{95.8} & 45.8 & 2.48 & 59.7 & \textbf{72.2} & 49.6\\
    \textbf{SheafPrune} & $80$ & 55.5 & 78.0 & 1111.07 & \textbf{83.4} & 79.7 & \textbf{69.6} & \textbf{79.0} & 73.8 & 69.8 & 63.3 & \textbf{69.2} & \textbf{92.0} & -- & -- & -- & -- & --\\
    \textbf{SheafPrune} & $90$ & 53.5 & 74.6 & 1082.23 & \textbf{81.1} & 77.2 & \textbf{53.6} & \textbf{62.7} & 59.4 & \textbf{68.1} & \textbf{61.1} & \textbf{63.0} & \textbf{84.6} & \textbf{29.4} & \textbf{3.87} & \textbf{74.2} & \textbf{71.4} & 63.0\\
    \midrule
    \multicolumn{19}{@{}l}{\textbf{LLaVA-1.5-7B}}\\
    \midrule
    No prune & $0$ & 58.7 & 75.8 & 1030.02 & 81.1 & 67.9 & 18.0 & 24.6 & 47.4 & 33.9 & 36.2 & 48.1 & -- & 39.7 & -- & -- & 72.3 & 0.0\\
    \addlinespace[1.2pt]
    Random & $10$ & 58.6 & 76.0 & 1031.50 & 80.1 & 68.4 & 17.7 & 23.7 & 47.1 & 33.6 & 36.2 & 47.8 & 99.2 & -- & -- & -- & -- & --\\
    Random & $30$ & 58.4 & 75.2 & 1026.39 & 78.9 & 67.4 & 17.4 & 21.3 & 43.6 & 33.4 & 35.5 & 46.8 & 96.7 & -- & -- & -- & -- & --\\
    Random & $50$ & 58.3 & 74.7 & 1029.63 & 79.0 & 67.4 & 16.1 & 19.7 & 40.9 & 33.8 & 35.4 & 46.2 & 94.7 & -- & -- & -- & -- & --\\
    Random & $60$ & 57.9 & 73.8 & \textbf{1029.04} & 77.5 & 66.8 & 15.5 & 17.3 & 39.3 & 33.8 & 36.0 & 45.4 & 92.8 & -- & -- & -- & -- & --\\
    Random & $70$ & 56.8 & 73.3 & 1004.01 & 77.3 & 65.6 & 14.3 & 15.8 & 35.3 & 33.2 & 35.0 & 44.2 & 89.5 & -- & -- & -- & -- & --\\
    Random & $80$ & 55.5 & 68.8 & 969.01 & 75.8 & 64.7 & 13.5 & 14.8 & 30.3 & 32.2 & 35.0 & 42.5 & 85.8 & -- & -- & -- & -- & --\\
    Random & $90$ & 51.5 & 62.5 & 890.70 & 73.5 & 61.2 & 11.9 & 10.6 & 22.9 & 30.9 & 32.9 & 38.9 & 77.6 & -- & -- & -- & -- & --\\
    \addlinespace[1.2pt]
    ToMe & $10$ & 58.7 & 76.0 & 1032.26 & 81.0 & 67.8 & 17.8 & 24.9 & 46.8 & 33.8 & \textbf{36.6} & 48.0 & 99.9 & -- & -- & -- & -- & --\\
    ToMe & $30$ & 58.5 & 74.8 & 1029.22 & 80.5 & 68.0 & 18.0 & 23.8 & 47.2 & 35.0 & 35.0 & 47.7 & 99.3 & -- & -- & -- & -- & --\\
    ToMe & $50$ & 58.5 & 74.5 & 1033.46 & 79.8 & 67.2 & 16.0 & 18.8 & 44.4 & 35.0 & 36.0 & 46.7 & 95.7 & 24.7 & 1.61 & 37.9 & 72.6 & 45.8\\
    ToMe & $60$ & 58.0 & 74.7 & 1025.11 & 79.3 & 66.6 & 15.8 & 19.3 & 44.4 & 35.1 & 36.6 & 46.6 & 95.7 & -- & -- & -- & -- & --\\
    ToMe & $70$ & 57.4 & 73.9 & 1006.63 & \textbf{79.7} & 66.2 & 16.0 & 17.8 & 42.7 & 34.4 & 35.6 & 46.0 & 93.9 & 19.6 & 2.03 & 50.7 & 72.0 & 63.9\\
    ToMe & $80$ & \textbf{57.3} & 72.5 & 997.93 & \textbf{78.6} & 64.9 & 14.5 & 15.0 & 35.7 & \textbf{35.9} & 33.1 & 44.3 & 89.6 & -- & -- & -- & -- & --\\
    ToMe & $90$ & \textbf{53.9} & 67.2 & 941.83 & \textbf{76.2} & 60.9 & 13.2 & 12.0 & 27.2 & 31.9 & 32.4 & 40.9 & 81.8 & 18.0 & 2.21 & 54.8 & 72.8 & 81.7\\
    \addlinespace[1.2pt]
    FastV & $10$ & 58.5 & 75.9 & \textbf{1036.93} & 80.9 & 68.0 & \textbf{18.2} & 24.6 & \textbf{47.6} & 33.9 & 35.5 & 48.0 & 100.0 & -- & -- & -- & -- & --\\
    FastV & $30$ & 58.1 & 74.9 & 1024.33 & 78.6 & 67.6 & 17.6 & 23.6 & 47.7 & 33.8 & 36.4 & 47.5 & 98.7 & -- & -- & -- & -- & --\\
    FastV & $50$ & 56.8 & 73.8 & 1013.15 & 77.3 & 67.1 & 17.1 & 21.6 & 45.2 & 33.1 & 36.9 & 46.5 & 96.4 & \textbf{24.4} & \textbf{1.63} & \textbf{38.7} & 57.7 & 45.8\\
    FastV & $60$ & 56.2 & 74.0 & 998.77 & 75.8 & 67.0 & 16.6 & 21.4 & 44.9 & 33.4 & 36.2 & 46.1 & 95.5 & -- & -- & -- & -- & --\\
    FastV & $70$ & 53.9 & 71.8 & 983.38 & 73.8 & 66.5 & 15.8 & 18.3 & 42.2 & 33.6 & 36.1 & 44.7 & 92.1 & 19.7 & 2.02 & 50.5 & 59.6 & 63.9\\
    FastV & $80$ & 52.1 & 69.3 & 942.30 & 65.7 & 65.1 & 14.8 & 14.5 & 38.2 & 32.8 & 35.2 & 42.1 & 86.4 & -- & -- & -- & -- & --\\
    FastV & $90$ & 44.0 & 54.2 & 810.13 & 27.6 & 54.0 & 13.0 & 9.4 & 23.7 & 30.1 & 31.4 & 31.6 & 67.5 & \textbf{17.3} & \textbf{2.30} & \textbf{56.5} & 59.9 & 81.8\\
    \addlinespace[1.2pt]
    SparseVLM & $10$ & \textbf{59.1} & 75.6 & 1034.65 & 80.9 & 67.8 & 18.2 & 24.4 & 47.3 & 33.6 & 35.9 & 48.0 & 99.8 & -- & -- & -- & -- & --\\
    SparseVLM & $30$ & 57.5 & 75.7 & 1033.31 & \textbf{81.5} & 68.0 & 18.1 & 24.7 & 46.8 & 33.5 & 35.2 & 47.8 & 99.4 & -- & -- & -- & -- & --\\
    SparseVLM & $50$ & 56.9 & 74.7 & \textbf{1035.86} & \textbf{81.3} & \textbf{67.9} & 17.4 & 22.8 & 43.8 & 33.4 & 35.8 & 47.1 & 97.5 & 29.2 & 1.36 & 26.5 & 58.2 & 46.1\\
    SparseVLM & $60$ & 56.2 & 72.3 & 1018.46 & 78.1 & 67.5 & 15.9 & 18.2 & 37.1 & 34.4 & 34.8 & 45.1 & 92.3 & -- & -- & -- & -- & --\\
    SparseVLM & $70$ & 49.8 & 64.6 & 972.47 & 72.9 & 65.6 & 13.7 & 10.6 & 21.8 & 35.2 & 34.1 & 40.3 & 81.4 & 27.2 & 1.46 & 31.5 & 58.5 & 64.0\\
    SparseVLM & $80$ & -- & -- & -- & -- & -- & -- & -- & -- & -- & -- & -- & -- & -- & -- & -- & -- & --\\
    SparseVLM & $90$ & -- & -- & -- & -- & -- & -- & -- & -- & -- & -- & -- & -- & -- & -- & -- & -- & --\\
    \addlinespace[1.2pt]
    PyramidDrop & $10$ & 58.4 & 75.7 & 1034.90 & 81.1 & 68.0 & 18.1 & 24.7 & 47.4 & 33.8 & 35.9 & 48.0 & 99.9 & -- & -- & -- & -- & --\\
    PyramidDrop & $30$ & 58.1 & 75.7 & 1029.69 & 80.8 & \textbf{68.0} & \textbf{18.1} & 24.7 & \textbf{47.8} & 33.4 & 35.5 & 47.9 & 99.6 & -- & -- & -- & -- & --\\
    PyramidDrop & $50$ & 57.8 & 75.8 & 1032.02 & 79.8 & 67.5 & \textbf{18.1} & 24.3 & \textbf{47.5} & 33.4 & 36.2 & 47.7 & 99.4 & 29.2 & 1.36 & 26.5 & 58.4 & \textbf{46.2}\\
    PyramidDrop & $60$ & 57.4 & 74.6 & 1022.74 & 78.4 & 67.4 & 17.5 & 23.4 & 46.5 & 34.8 & 34.5 & 47.1 & 97.9 & -- & -- & -- & -- & --\\
    PyramidDrop & $70$ & 55.3 & 72.1 & 977.35 & 73.5 & 65.6 & 16.0 & 17.1 & 40.4 & 33.1 & 34.4 & 44.2 & 90.7 & 26.5 & 1.50 & 33.3 & 58.5 & \textbf{64.1}\\
    PyramidDrop & $80$ & -- & -- & -- & -- & -- & -- & -- & -- & -- & -- & -- & -- & -- & -- & -- & -- & --\\
    PyramidDrop & $90$ & -- & -- & -- & -- & -- & -- & -- & -- & -- & -- & -- & -- & -- & -- & -- & -- & --\\
    \addlinespace[1.2pt]
    G-Prune & $10$ & 58.2 & \textbf{76.1} & 1025.32 & 80.7 & \textbf{68.5} & 18.0 & 24.6 & 47.3 & 33.4 & 35.1 & 47.9 & 99.4 & -- & -- & -- & -- & --\\
    G-Prune & $30$ & 57.7 & 75.3 & 1015.39 & 79.7 & 66.9 & 17.5 & 24.7 & 47.7 & 35.1 & 35.8 & 47.7 & 99.3 & -- & -- & -- & -- & --\\
    G-Prune & $50$ & 56.6 & 73.9 & 1017.94 & 77.9 & 67.3 & 17.4 & 24.4 & 47.2 & 34.1 & 36.4 & 47.2 & 98.4 & 24.7 & 1.61 & 37.9 & 73.6 & 45.8\\
    G-Prune & $60$ & 56.7 & 73.0 & 1021.16 & 76.9 & 66.7 & 17.6 & 24.0 & 46.7 & 34.0 & 36.2 & 46.8 & 97.9 & -- & -- & -- & -- & --\\
    G-Prune & $70$ & 55.1 & 72.9 & 1013.33 & 75.5 & 65.2 & 17.1 & 22.6 & 46.5 & 33.1 & 36.1 & 46.0 & 96.0 & \textbf{19.0} & \textbf{2.09} & \textbf{52.2} & 73.6 & 63.9\\
    G-Prune & $80$ & 52.7 & 71.0 & 978.89 & 71.5 & 64.6 & 16.4 & 22.0 & 44.9 & 33.9 & 35.6 & 44.8 & 93.5 & -- & -- & -- & -- & --\\
    G-Prune & $90$ & 47.3 & 62.8 & 865.14 & 52.7 & 57.9 & 14.6 & 16.8 & 40.2 & 33.2 & 34.2 & 39.1 & 82.4 & 18.1 & 2.19 & 54.3 & 74.8 & 81.7\\
    \addlinespace[1.2pt]
    FasterVLM & $10$ & 58.8 & 75.9 & 1025.69 & 81.4 & 68.0 & 18.0 & 24.9 & 47.2 & 33.6 & 36.0 & 48.0 & 99.9 & -- & -- & -- & -- & --\\
    FasterVLM & $30$ & \textbf{58.9} & \textbf{76.1} & 1034.64 & 80.9 & 67.3 & 17.4 & 25.2 & 46.7 & 35.0 & 37.1 & 48.1 & 100.3 & -- & -- & -- & -- & --\\
    FasterVLM & $50$ & 58.5 & \textbf{76.0} & 1024.47 & 80.1 & 67.4 & 17.4 & \textbf{24.8} & 46.5 & 35.6 & 36.0 & 47.9 & 99.7 & 29.4 & 1.35 & 25.9 & 73.6 & 45.8\\
    FasterVLM & $60$ & 58.2 & 75.0 & 1022.66 & 78.2 & 67.5 & 17.4 & 23.7 & 46.1 & \textbf{35.9} & 36.2 & 47.5 & 98.9 & -- & -- & -- & -- & --\\
    FasterVLM & $70$ & 55.9 & 74.3 & 1023.51 & 76.4 & 66.8 & 17.6 & 22.7 & 45.5 & \textbf{35.8} & 37.1 & 46.9 & 97.9 & 23.8 & 1.67 & 40.1 & \textbf{73.9} & 63.9\\
    FasterVLM & $80$ & 55.2 & 72.0 & 1003.94 & 73.2 & 66.0 & 17.0 & 22.0 & 44.9 & 34.9 & 36.2 & 45.7 & 95.5 & -- & -- & -- & -- & --\\
    FasterVLM & $90$ & 51.7 & 68.9 & \textbf{974.25} & 65.5 & 64.9 & 16.1 & 18.3 & 41.7 & \textbf{34.5} & \textbf{35.5} & 43.2 & \textbf{90.1} & 21.5 & 1.85 & 45.9 & \textbf{74.8} & 81.7\\
    \addlinespace[1.2pt]
    VisionZip & $10$ & 58.7 & 75.9 & 1023.18 & \textbf{81.4} & 68.0 & 17.9 & \textbf{25.1} & 47.2 & 33.8 & 36.2 & \textbf{48.1} & \textbf{100.0} & -- & -- & -- & -- & --\\
    VisionZip & $30$ & 58.6 & 75.9 & 1034.44 & 80.9 & 67.3 & 17.3 & \textbf{25.3} & 46.7 & \textbf{35.2} & \textbf{37.4} & \textbf{48.2} & \textbf{100.3} & -- & -- & -- & -- & --\\
    VisionZip & $50$ & \textbf{58.8} & 75.6 & 1019.21 & 79.5 & 67.5 & 17.4 & 24.5 & 46.3 & \textbf{35.6} & 36.8 & 47.8 & 99.6 & 33.4 & 1.19 & 16.0 & 73.6 & 45.8\\
    VisionZip & $60$ & 57.6 & 74.6 & 1024.62 & 79.0 & 67.2 & 17.7 & 23.5 & 46.0 & 35.4 & 36.6 & 47.4 & 98.8 & -- & -- & -- & -- & --\\
    VisionZip & $70$ & 56.5 & 74.4 & \textbf{1025.46} & 76.9 & 67.0 & 17.7 & 23.1 & 45.6 & 35.1 & \textbf{37.1} & 47.0 & 98.1 & 26.5 & 1.50 & 33.3 & 72.6 & 63.9\\
    VisionZip & $80$ & 54.6 & 72.6 & \textbf{1004.97} & 74.5 & 66.1 & 16.8 & 21.6 & 44.4 & 34.1 & 35.1 & 45.6 & 94.8 & -- & -- & -- & -- & --\\
    VisionZip & $90$ & 51.9 & 69.5 & 969.49 & 68.0 & \textbf{65.1} & \textbf{16.1} & 17.3 & 40.8 & 34.4 & 34.4 & 43.2 & 89.6 & 22.4 & 1.77 & 43.5 & 74.3 & 81.7\\
    \addlinespace[1.2pt]
    PruMerge & $10$ & 58.7 & 75.9 & 1024.12 & 81.3 & 68.0 & 17.8 & 25.0 & 47.2 & 33.6 & 36.1 & 48.0 & 99.9 & -- & -- & -- & -- & --\\
    PruMerge & $30$ & 58.7 & 76.0 & \textbf{1035.17} & 80.5 & 67.3 & 17.1 & 25.0 & 47.0 & 35.0 & 37.0 & 48.1 & 100.0 & -- & -- & -- & -- & --\\
    PruMerge & $50$ & 58.6 & 75.6 & 1026.63 & 80.2 & 67.3 & 17.5 & 24.0 & 46.5 & 35.4 & \textbf{37.2} & 47.9 & 99.7 & 40.5 & 0.98 & -2.0 & \textbf{73.7} & 45.8\\
    PruMerge & $60$ & 57.7 & 75.1 & 1025.16 & 78.9 & 67.5 & 17.6 & 24.1 & 46.9 & 35.0 & \textbf{37.5} & 47.7 & 99.5 & -- & -- & -- & -- & --\\
    PruMerge & $70$ & 56.7 & 74.4 & 1023.79 & 77.7 & 66.7 & 17.0 & 22.5 & 45.8 & 34.8 & 36.8 & 46.9 & 97.4 & 34.5 & 1.15 & 13.0 & 72.9 & 63.9\\
    PruMerge & $80$ & 54.8 & \textbf{73.5} & 1003.64 & 77.3 & \textbf{66.6} & 16.2 & 20.4 & 44.2 & 33.2 & \textbf{36.2} & 45.8 & 94.5 & -- & -- & -- & -- & --\\
    PruMerge & $90$ & 53.3 & \textbf{70.1} & 964.23 & 72.2 & 64.8 & 15.3 & 16.3 & 40.3 & 33.0 & 34.6 & \textbf{43.4} & 89.0 & 27.0 & 1.47 & 32.0 & 74.4 & 81.7\\
    \addlinespace[1.2pt]
    \textbf{SheafPrune} & $10$ & 58.7 & 75.8 & 1029.62 & 81.1 & 67.9 & 18.0 & 24.6 & 47.4 & \textbf{33.9} & 36.2 & 48.0 & 100.0 & -- & -- & -- & -- & --\\
    \textbf{SheafPrune} & $30$ & 58.7 & 75.8 & 1029.65 & 81.1 & 67.8 & 18.0 & 24.6 & 47.4 & 33.9 & 36.2 & 48.0 & 100.0 & -- & -- & -- & -- & --\\
    \textbf{SheafPrune} & $50$ & 58.6 & 75.7 & 1028.76 & 81.0 & 67.8 & 18.0 & 24.6 & 47.4 & 33.9 & 36.2 & \textbf{48.0} & \textbf{99.9} & 26.8 & 1.48 & 32.4 & 73.4 & 45.8\\
    \textbf{SheafPrune} & $60$ & \textbf{58.4} & \textbf{75.5} & 1026.86 & \textbf{80.7} & \textbf{67.8} & \textbf{18.0} & \textbf{24.6} & \textbf{47.4} & 33.9 & 36.2 & \textbf{47.9} & \textbf{99.8} & -- & -- & -- & -- & --\\
    \textbf{SheafPrune} & $70$ & \textbf{57.7} & \textbf{74.9} & 1019.97 & 79.3 & \textbf{67.3} & \textbf{17.9} & \textbf{24.2} & \textbf{47.1} & 33.9 & 36.2 & \textbf{47.5} & \textbf{99.0} & 21.2 & 1.87 & 46.5 & 72.9 & 64.0\\
    \textbf{SheafPrune} & $80$ & 56.2 & 73.2 & 995.12 & 76.5 & 66.1 & \textbf{17.3} & \textbf{23.5} & \textbf{46.6} & 33.6 & 35.9 & \textbf{46.4} & \textbf{96.8} & -- & -- & -- & -- & --\\
    \textbf{SheafPrune} & $90$ & 51.4 & 67.6 & 924.23 & 63.6 & 61.6 & 15.8 & \textbf{19.7} & \textbf{42.6} & 33.9 & 35.1 & 42.4 & 89.0 & 18.8 & 2.11 & 52.6 & 74.6 & \textbf{81.9}\\
    \midrule
    \multicolumn{19}{@{}l}{\textbf{InternVL3-8B}}\\
    \midrule
    No prune & $0$ & 61.3 & 79.3 & 1253.19 & 89.6 & 87.0 & 86.5 & 92.5 & 82.3 & 70.0 & 61.9 & 75.5 & -- & 128.7 & -- & -- & 70.9 & 0.0\\
    \addlinespace[1.2pt]
    Random & $10$ & 61.2 & 79.1 & 1246.94 & \textbf{89.9} & 87.2 & 85.9 & 91.7 & 81.7 & \textbf{70.2} & 61.8 & 75.3 & 99.7 & -- & -- & -- & -- & --\\
    Random & $30$ & 61.4 & 79.1 & 1235.22 & 89.4 & 86.9 & 84.0 & 86.4 & 79.4 & 68.9 & 60.6 & 74.0 & 98.1 & -- & -- & -- & -- & --\\
    Random & $50$ & \textbf{61.7} & 78.8 & 1223.11 & 89.0 & 86.6 & 77.8 & 75.0 & 74.6 & 67.1 & 58.5 & 71.3 & 94.8 & -- & -- & -- & -- & --\\
    Random & $60$ & 60.9 & 78.6 & 1202.92 & 89.1 & 86.6 & 72.9 & 64.2 & 69.8 & 65.6 & 59.1 & 69.0 & 92.1 & -- & -- & -- & -- & --\\
    Random & $70$ & 61.0 & 77.6 & 1195.06 & 88.8 & 85.6 & 64.3 & 51.6 & 62.6 & 64.5 & 58.2 & 65.7 & 88.3 & -- & -- & -- & -- & --\\
    Random & $80$ & 59.3 & 74.9 & 1170.54 & 88.3 & 84.2 & 51.4 & 39.7 & 54.5 & 62.0 & 54.2 & 61.0 & 82.5 & -- & -- & -- & -- & --\\
    Random & $90$ & \textbf{56.8} & \textbf{70.0} & 1103.52 & \textbf{85.6} & \textbf{80.0} & 35.0 & 25.4 & 39.7 & 57.0 & 50.5 & 53.9 & 73.6 & -- & -- & -- & -- & --\\
    \addlinespace[1.2pt]
    ToMe & $10$ & 61.1 & 79.4 & 1251.28 & 89.3 & 87.2 & 85.7 & 91.9 & 81.6 & 69.4 & 61.5 & 75.2 & 99.5 & -- & -- & -- & -- & --\\
    ToMe & $30$ & 61.3 & \textbf{79.8} & 1253.62 & 89.6 & 86.5 & 84.8 & 89.0 & 80.1 & 69.6 & 61.6 & 74.7 & 99.1 & -- & -- & -- & -- & --\\
    ToMe & $50$ & 60.3 & 79.2 & 1228.97 & 89.1 & 86.1 & 78.8 & 79.5 & 77.8 & 69.5 & 58.1 & 72.2 & 95.9 & \textbf{71.5} & \textbf{1.80} & \textbf{44.4} & 70.6 & 42.9\\
    ToMe & $60$ & 60.5 & \textbf{79.3} & 1226.28 & 88.5 & 86.2 & 75.5 & 74.9 & 76.0 & 68.5 & 57.5 & 71.1 & 94.6 & -- & -- & -- & -- & --\\
    ToMe & $70$ & 60.5 & 77.8 & 1199.65 & 88.6 & 85.8 & 63.6 & 62.1 & 72.4 & 67.4 & 57.4 & 67.8 & 90.8 & \textbf{46.3} & \textbf{2.78} & \textbf{64.0} & 71.3 & 59.3\\
    ToMe & $80$ & 59.7 & 75.2 & 1184.32 & 88.1 & 83.7 & 46.3 & 44.4 & 63.8 & 66.8 & 57.0 & 62.7 & 84.8 & -- & -- & -- & -- & --\\
    ToMe & $90$ & 55.3 & 69.2 & 1105.18 & 84.9 & 78.3 & 27.0 & 23.2 & 47.3 & 63.0 & 51.6 & 53.9 & 73.8 & 33.5 & 3.84 & 74.0 & \textbf{72.0} & 75.3\\
    \addlinespace[1.2pt]
    FastV & $10$ & 61.3 & 79.3 & 1250.62 & 89.6 & 87.1 & 86.4 & 92.1 & 82.2 & 69.9 & \textbf{61.9} & 75.5 & 99.9 & -- & -- & -- & -- & --\\
    FastV & $30$ & 61.2 & 79.3 & 1252.63 & 89.6 & 87.0 & 86.0 & 91.8 & 82.5 & 69.8 & 61.0 & 75.3 & 99.7 & -- & -- & -- & -- & --\\
    FastV & $50$ & 60.8 & \textbf{79.5} & 1252.67 & 88.6 & \textbf{87.5} & 84.8 & 89.2 & \textbf{82.3} & 70.6 & 61.6 & 75.0 & 99.4 & 74.8 & 1.72 & 41.9 & 54.8 & 43.0\\
    FastV & $60$ & 60.9 & 79.2 & 1245.49 & 88.8 & \textbf{87.2} & 82.5 & 85.3 & \textbf{81.6} & 70.1 & 61.2 & 74.1 & 98.4 & -- & -- & -- & -- & --\\
    FastV & $70$ & 60.5 & \textbf{79.1} & \textbf{1240.14} & 88.7 & \textbf{86.2} & 76.5 & 75.5 & \textbf{81.4} & 70.4 & 61.5 & 72.4 & 96.4 & 52.3 & 2.46 & 59.3 & 54.7 & 59.4\\
    FastV & $80$ & 59.5 & \textbf{78.1} & \textbf{1218.23} & 87.6 & \textbf{84.9} & 62.6 & 60.4 & \textbf{75.6} & 66.8 & 58.1 & 67.7 & 90.7 & -- & -- & -- & -- & --\\
    FastV & $90$ & 52.4 & 67.8 & 1093.68 & 79.8 & 72.0 & 18.7 & 21.2 & 45.6 & 50.9 & 45.2 & 49.3 & 67.6 & 34.2 & 3.76 & 73.4 & 54.2 & 75.3\\
    \addlinespace[1.2pt]
    SparseVLM & $10$ & 61.4 & \textbf{79.4} & 1252.88 & 89.3 & 87.0 & 86.2 & 92.5 & \textbf{82.5} & 70.0 & 61.4 & 75.4 & 99.9 & -- & -- & -- & -- & --\\
    SparseVLM & $30$ & \textbf{61.5} & 79.1 & 1249.17 & 89.4 & \textbf{87.4} & 86.0 & 91.8 & 82.3 & 69.9 & 61.5 & 75.3 & 99.8 & -- & -- & -- & -- & --\\
    SparseVLM & $50$ & 61.0 & 78.6 & \textbf{1252.75} & 88.8 & 86.9 & 82.6 & 88.4 & 81.2 & 70.2 & 61.9 & 74.4 & 98.8 & 77.5 & 1.66 & 39.8 & 54.1 & 43.9\\
    SparseVLM & $60$ & 59.9 & 78.2 & \textbf{1245.86} & 88.5 & 86.9 & 75.4 & 83.4 & 78.3 & 70.6 & 60.8 & 72.7 & 96.6 & -- & -- & -- & -- & --\\
    SparseVLM & $70$ & 58.3 & 76.4 & 1222.94 & 87.5 & 84.9 & 45.2 & 60.8 & 66.5 & 68.8 & 61.2 & 65.3 & 88.0 & 59.3 & 2.17 & 53.9 & 54.3 & 60.0\\
    SparseVLM & $80$ & -- & -- & -- & -- & -- & -- & -- & -- & -- & -- & -- & -- & -- & -- & -- & -- & --\\
    SparseVLM & $90$ & -- & -- & -- & -- & -- & -- & -- & -- & -- & -- & -- & -- & -- & -- & -- & -- & --\\
    \addlinespace[1.2pt]
    PyramidDrop & $10$ & \textbf{61.5} & 78.9 & 1252.95 & 89.4 & \textbf{87.3} & 86.3 & 92.3 & 82.3 & 69.8 & 61.8 & 75.4 & 99.9 & -- & -- & -- & -- & --\\
    PyramidDrop & $30$ & 61.3 & 79.2 & \textbf{1256.13} & 89.3 & 87.0 & 86.3 & 92.2 & 82.4 & 70.1 & 61.1 & 75.4 & 99.8 & -- & -- & -- & -- & --\\
    PyramidDrop & $50$ & 61.1 & 78.7 & 1241.31 & 88.7 & 86.5 & 85.0 & 89.8 & 81.9 & 70.1 & 61.8 & 74.8 & 99.1 & 76.2 & 1.69 & 40.8 & 53.8 & \textbf{44.4}\\
    PyramidDrop & $60$ & 60.3 & 78.0 & 1230.00 & 88.5 & 85.9 & 81.5 & 85.9 & 80.9 & 70.0 & 62.8 & 73.8 & 97.9 & -- & -- & -- & -- & --\\
    PyramidDrop & $70$ & 58.7 & 76.2 & 1160.00 & 86.8 & 81.3 & 60.6 & 62.4 & 68.8 & 64.5 & 57.5 & 65.8 & 88.1 & 59.3 & 2.17 & 53.9 & 53.7 & \textbf{60.1}\\
    PyramidDrop & $80$ & -- & -- & -- & -- & -- & -- & -- & -- & -- & -- & -- & -- & -- & -- & -- & -- & --\\
    PyramidDrop & $90$ & -- & -- & -- & -- & -- & -- & -- & -- & -- & -- & -- & -- & -- & -- & -- & -- & --\\
    \addlinespace[1.2pt]
    G-Prune & $10$ & 61.3 & 79.3 & 1249.82 & 89.5 & 87.2 & 86.2 & \textbf{92.9} & 82.4 & 69.5 & 61.4 & 75.4 & 99.8 & -- & -- & -- & -- & --\\
    G-Prune & $30$ & 61.0 & 79.2 & 1246.37 & \textbf{89.7} & 86.5 & 85.7 & 91.8 & \textbf{82.6} & \textbf{70.1} & 61.5 & 75.3 & 99.7 & -- & -- & -- & -- & --\\
    G-Prune & $50$ & 61.3 & 79.3 & 1247.56 & 89.4 & 86.6 & 81.9 & 88.0 & 81.5 & \textbf{70.9} & \textbf{62.9} & 74.6 & 99.0 & 72.3 & 1.78 & 43.8 & 70.8 & 42.9\\
    G-Prune & $60$ & 60.5 & 79.2 & 1242.41 & 89.2 & 86.1 & 78.2 & 84.3 & 81.0 & \textbf{70.8} & \textbf{62.8} & 73.6 & 97.8 & -- & -- & -- & -- & --\\
    G-Prune & $70$ & 59.7 & 78.0 & 1233.63 & 89.3 & 85.5 & 72.2 & 77.8 & 78.5 & \textbf{71.0} & \textbf{63.2} & 71.9 & 95.8 & 47.0 & 2.74 & 63.5 & 71.0 & 59.3\\
    G-Prune & $80$ & 56.9 & 76.5 & 1188.04 & 88.0 & 83.3 & 57.0 & 66.0 & 72.7 & \textbf{70.4} & \textbf{64.2} & 67.8 & 90.8 & -- & -- & -- & -- & --\\
    G-Prune & $90$ & 52.9 & 69.3 & \textbf{1106.53} & 84.7 & 78.0 & 31.0 & \textbf{42.8} & \textbf{56.8} & \textbf{70.2} & \textbf{59.4} & \textbf{58.5} & \textbf{79.4} & \textbf{30.2} & \textbf{4.26} & \textbf{76.5} & 71.4 & 75.2\\
    \addlinespace[1.2pt]
    \textbf{SheafPrune} & $10$ & 61.3 & 79.3 & \textbf{1252.99} & 89.6 & 87.0 & \textbf{86.5} & 92.5 & 82.3 & 69.9 & 61.8 & \textbf{75.5} & \textbf{100.0} & -- & -- & -- & -- & --\\
    \textbf{SheafPrune} & $30$ & 61.3 & 79.3 & 1251.85 & 89.6 & 87.0 & \textbf{86.4} & \textbf{92.3} & 82.2 & 69.9 & \textbf{61.7} & \textbf{75.4} & \textbf{99.9} & -- & -- & -- & -- & --\\
    \textbf{SheafPrune} & $50$ & 61.3 & 79.2 & 1247.33 & \textbf{89.5} & 86.8 & \textbf{85.9} & \textbf{91.0} & 81.8 & 70.0 & 61.9 & \textbf{75.2} & \textbf{99.6} & 76.2 & 1.69 & 40.8 & \textbf{71.9} & 42.9\\
    \textbf{SheafPrune} & $60$ & \textbf{61.2} & 79.0 & 1237.98 & \textbf{89.5} & 86.5 & \textbf{84.5} & \textbf{88.4} & 81.0 & 70.0 & 61.8 & \textbf{74.6} & \textbf{98.9} & -- & -- & -- & -- & --\\
    \textbf{SheafPrune} & $70$ & \textbf{61.0} & 78.0 & 1213.21 & \textbf{89.5} & 86.0 & \textbf{81.6} & \textbf{83.3} & 79.1 & 70.0 & 61.6 & \textbf{73.4} & \textbf{97.3} & 50.3 & 2.56 & 60.9 & \textbf{72.3} & 59.6\\
    \textbf{SheafPrune} & $80$ & \textbf{59.9} & 75.8 & 1186.28 & \textbf{89.2} & 84.0 & \textbf{71.2} & \textbf{69.8} & 71.3 & 68.1 & 59.6 & \textbf{69.1} & \textbf{92.2} & -- & -- & -- & -- & --\\
    \textbf{SheafPrune} & $90$ & 54.6 & 67.3 & 1074.99 & 85.1 & 76.9 & \textbf{41.2} & 38.9 & 42.2 & 63.6 & 55.9 & 56.4 & 76.5 & 30.4 & 4.24 & 76.4 & 71.6 & \textbf{75.4}\\
\end{longtable}
\endgroup

\section{Statistical Scope of the Final Results}
\label{app:significance}

The reported tables contain aggregate benchmark scores rather than the paired predictions and repeated-training results required for inferential comparisons. They support comparisons of the displayed point estimates, but not paired bootstrap tests, McNemar tests, seed-variance estimates or multiplicity-corrected significance claims. No universal accuracy ``noise floor'' is inferred from these aggregates. A future statistical comparison should fix the hyperparameter policy, retain the corresponding per-sample predictions and distinguish evaluation uncertainty from variation across training seeds.

\section{Component Ablation}
\label{app:component_ablation}

Table~\ref{tab:app_ablation_full} expands
Table~\ref{tab:exp_ablation} with initialization variants and task groups.
The full method improves over uniform initialization by $0.93$, $1.25$
and $1.20$ RA points at $p=0.5,0.7,0.9$, respectively.
At $p=0.9$, removing relevance ($\lambda=0$) gives $86.40\%$ RA
and removing the visual prior ($\lambda=1$) gives $85.00\%$,
both below the uniform arm's $88.19\%$ and the full method's $89.39\%$.
These comparisons support the added value of combining relevance
with the visual prior within the fixed graph operator,
while uniform initialization remains a strong baseline.
The Text-group difference between the full method and uniform
initialization is $1.65$ points.

\textbf{Dense operator definition.} The documented dense variant replaces the off-support logits by zero before row softmax. With $M_{ij}=\mathbb1[j\in\mathcal N(i)]$ on valid pairs, its mathematical form is
\begin{equation}
T^{\mathrm{dense}}_{ij}=\frac{\exp(M_{ij}\operatorname{sim}_{ij})}{\sum_{l:v_l=1}\exp(M_{il}\operatorname{sim}_{il})}.
\label{eq:app_dense_variant}
\end{equation}
Every valid non-edge then has positive weight because $\exp(0)=1$. This changes both propagation support and normalization; it is not a denominator-only intervention. The reported ablation keeps the threshold-graph degree correction at $\delta=1$. It is an internal SheafPrune intervention, not a claim that official G-Prune omits masking.

Table~\ref{tab:app_2x2} reports a separate norm-plus-graph family on LLaVA-1.5. Its uniform and norm initializations differ by at most $0.1$ points with propagation, but these are not SheafPrune arms. They provide a related observation rather than a replication of the present ablation.

\begin{table}[!htbp]
    \caption{\textbf{Complete component ablation of SheafPrune on LLaVA-NeXT-8B.} The default configuration matches Table~\ref{tab:exp_ablation}; ``Text'' averages ChartQA, DocVQA and TextVQA,
    ``OCR'' adds HR-Bench 4K and 8K, and ``Gen.'' averages GQA, VQAv2, MMBench-EN, POPE and MME.
    Rows marked $\dagger$ give additional initialization variants.}
    \label{tab:app_ablation_full}
    \centering
    \small
    \setlength{\tabcolsep}{4pt}
    \begin{tabular}{@{}l cccc ccc ccc@{}}
        \toprule
        & \multicolumn{4}{c}{Configuration} & \multicolumn{3}{c}{Ten-benchmark RA (\%)} & \multicolumn{3}{c}{At $p{=}0.9$}\\
        \cmidrule(lr){2-5}\cmidrule(lr){6-8}\cmidrule(lr){9-11}
        Arm & $s^{(0)}$ & Prop. & Norm. & $\delta$ & $0.5$ & $0.7$ & $0.9$ & Text & OCR & Gen.\\
        \midrule
        Random selection & --- & $\times$ & --- & --- & 92.12 & 86.32 & 73.38 & 39.20 & 54.98 & 91.77\\
        Degree only & uniform & $\times$ & --- & $1$ & 95.77 & 92.14 & 81.38 & 57.09 & 69.03 & 93.74\\
        Visual prior only & $\overline n$ & $\times$ & --- & $0$ & 91.13 & 84.40 & 70.23 & 34.10 & 49.88 & 90.57\\
        Relevance only$^\dagger$ & $\widehat r$ & $\times$ & --- & $0$ & 91.81 & 85.80 & 72.58 & 38.00 & 53.78 & 91.37\\
        \midrule
        Uniform initial score & uniform & $\checkmark$ & support & $1$ & 98.50 & 96.72 & 88.19 & 73.32 & 81.54 & 94.63\\
        No relevance ($\lambda{=}0$)$^\dagger$ & $\overline n$ & $\checkmark$ & support & $1$ & 98.49 & 96.11 & 86.40 & 69.17 & 77.28 & 95.53\\
        No visual prior ($\lambda{=}1$)$^\dagger$ & $\widehat r$ & $\checkmark$ & support & $1$ & 98.35 & 95.19 & 85.00 & 66.97 & 75.08 & 94.93\\
        \midrule
        No degree correction & both & $\checkmark$ & support & $0$ & 91.88 & 85.97 & 72.80 & 38.30 & 54.08 & 91.52\\
        Dense normalization & both & $\checkmark$ & dense & $1$ & 91.74 & 85.78 & 72.68 & 38.10 & 53.88 & 91.47\\
        \midrule
        \textbf{SheafPrune} & both & $\checkmark$ & support & $1$ & \textbf{99.43} & \textbf{97.97} & \textbf{89.39} & \textbf{74.97} & \textbf{83.08} & \textbf{95.71}\\
        \bottomrule
    \end{tabular}
\end{table}

\begin{table}[!htbp]
    \caption{Component ablation of the \emph{norm-plus-graph baseline family} on LLaVA-1.5-7B: accuracy retention (\%) averaged over ten benchmarks. These are arms of that family, not of SheafPrune. B = random selection, N = feature norm without a graph, U = uniform initialization with diffusion, F = norm initialization with diffusion.}
    \label{tab:app_2x2}
    \centering
    \small
    \begin{tabular}{@{}lccccccc@{}}
        \toprule
        Configuration & $\rho{=}0.9$ & $0.7$ & $0.5$ & $0.4$ & $0.3$ & $0.2$ & $0.1$\\
        \midrule
        B: no norm, no graph & 99.24 & 96.66 & 94.71 & 92.75 & 89.54 & 85.85 & 77.60\\
        N: norm, no graph & 99.52 & 97.00 & 89.38 & 83.23 & 74.62 & 59.48 & 53.72\\
        U: uniform, graph & 99.41 & 99.38 & 98.40 & 97.88 & 96.06 & 93.47 & 82.39\\
        F: norm, graph & 99.43 & 99.30 & 98.43 & 97.88 & 95.96 & 93.48 & 82.39\\
        \midrule
        F $-$ U (value of the norm prior) & $+0.02$ & $-0.07$ & $+0.03$ & $-0.00$ & $-0.10$ & $+0.01$ & $-0.00$\\
        U $-$ B (value of the diffusion) & $+0.17$ & $+2.71$ & $+3.69$ & $+5.12$ & $+6.52$ & $+7.62$ & $+4.79$\\
        \bottomrule
    \end{tabular}
\end{table}

\subsection{Hyperparameter Analysis}
\label{app:hyper}

The component study above already covers the remaining budgets, so it is not repeated here; Table~\ref{tab:app_hyper_sensitivity} reports the remaining two sensitivity studies.

\begin{table}[!htbp]
    \caption{Sensitivity of the propagation depth $t$ and of the similarity threshold $s$, measured as the mean score over four benchmarks at $\downarrow90\%$ on Qwen2.5-VL. These are point estimates, without a measured noise threshold.}
    \label{tab:app_hyper_sensitivity}
    \centering
    \small
    \setlength{\tabcolsep}{5pt}
    \begin{tabular}{@{}cc@{\hskip 24pt}cc@{}}
        \toprule
        \multicolumn{2}{c}{\emph{(a) Propagation depth}} & \multicolumn{2}{c}{\emph{(b) Similarity threshold}}\\
        \cmidrule(lr){1-2}\cmidrule(lr){3-4}
        $t$ & Mean & $s$ & Mean\\
        \midrule
        $1$ & 54.06 & $0.3$ & 49.11\\
        $3$ & \textbf{54.23} & $0.4$ & 53.61\\
        $5$ & 53.85 & $0.5$ & \textbf{53.85}\\
        $10$ & 53.85 & $0.6$ & 51.18\\
        $20$ & 53.06 & $0.7$ & 44.90\\
        \bottomrule
    \end{tabular}
\end{table}

\textbf{Graph construction and hyperparameters.}
For $t\in\{1,3,5,10\}$ the reported mean scores range from $53.85$ to $54.23$, a spread of $0.38$ points; $t=20$ gives $53.06$. This describes the tested configuration and does not establish a noise threshold or eliminate the need for depth selection elsewhere. The separate threshold sweep uses an absolute cosine threshold $s$, rather than the default retained-relation fraction $q_v=0.05$; its largest displayed mean is $53.85$ at $s=0.5$. The two sweeps should not be interpreted as one common configuration grid without their run settings.

The existing $k$-NN comparison reports a $23$--$24$-point decrease in the DocVQA score for $k\in\{16,32,64,128\}$. It changes neighborhood construction and may change symmetry and degree distribution together. It therefore supports sensitivity to the tested graph construction, without isolating the effect of degree information. Complete per-$k$ run scores and the symmetrization convention are needed for a more specific attribution.

\section{Spatial Coverage Diagnosis}
\label{app:coverage}

This section states how the diagnosis of Section~\ref{sec:exp_ablation} is measured and reports the coverage figures it is based on (Table~\ref{tab:exp_coverage}). Retained tokens are mapped back onto the image grid, and two quantities are recorded on the original visual-token layout so that benchmarks with different visual sequence lengths remain comparable: grid coverage, the fraction of $8\times8$ cells that keep at least one token, and the mean nearest-neighbor distance between retained tokens. The retained subsets show similar spatial concentration for both graph-based selectors: at $\rho=0.1$ on GQA the coverage is $0.417$ for SheafPrune and $0.385$ for G-Prune, against $0.500$ for random selection. An explicit coverage constraint is a possible future extension, but its effect is not tested here. These cross-task observations do not isolate coverage as the cause of accuracy differences, because token counts, image content and tokenization also change.

\begin{table}[!htbp]
    \centering
    \small
    \caption{Grid coverage at $\downarrow90\%$: fraction of $8\times8$ cells that keep a token.}
    \label{tab:exp_coverage}
    \setlength{\tabcolsep}{4.5pt}
    \begin{tabular}{@{}lccc@{}}
        \toprule
        & $N_v$ & Random & Ours\\
        \midrule
        GQA & $345$ & $0.500$ & $0.417$\\
        VQAv2 & $345$ & $0.432$ & $0.333$\\
        DocVQA & $4032$ & $0.859$ & $0.771$\\
        HR-Bench & $4096$ & $1.000$ & $0.844$\\
        \bottomrule
    \end{tabular}
\end{table}

\section{Additional Qualitative Samples}
\label{app:qualitative}

\begin{figure}[H]
    \centering
    \includegraphics[
        width=0.9\textwidth,
        height=0.7\textheight,
        keepaspectratio
    ]{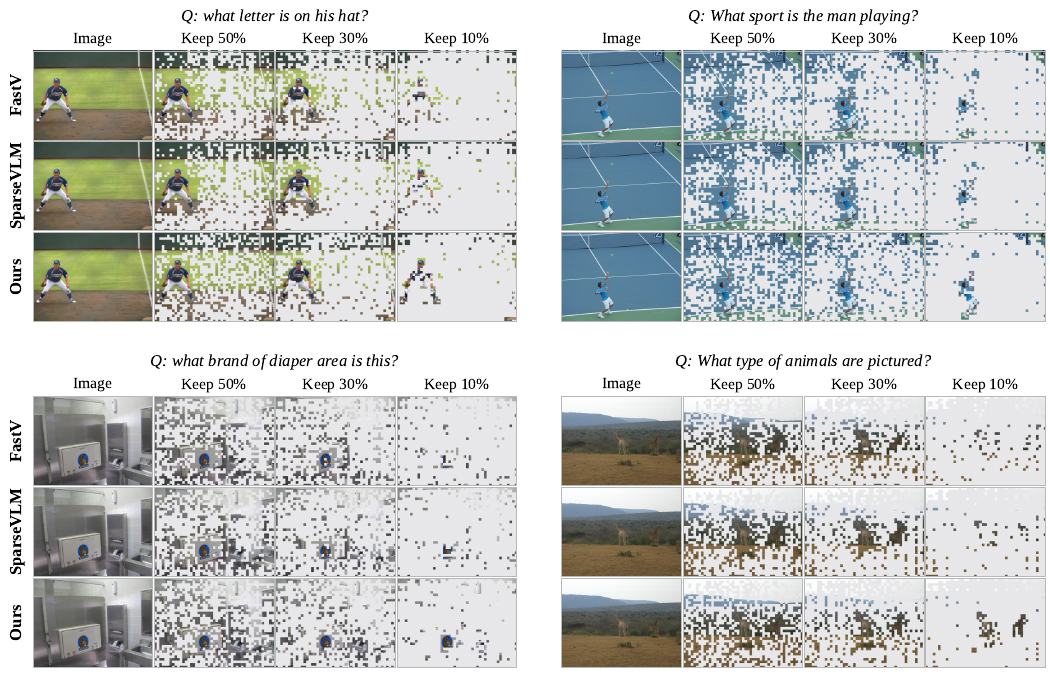}
    \caption{\textbf{Additional qualitative samples.}
    Same protocol as Figure~\ref{fig:qualitative}: FastV,
    SparseVLM and SheafPrune at $\rho\in\{0.5,0.3,0.1\}$
    on LLaVA-NeXT-8B, with pruned positions greyed out.}
    \label{fig:qualitative_appendix}
\end{figure}

Figure~\ref{fig:qualitative_appendix} repeats the protocol of
Figure~\ref{fig:qualitative} on four further samples, two from
TextVQA and two from VQAv2, shown at the same three budgets.

\section{Local Operator Visualizations}
\label{app:operator_examples}
Figure~\ref{fig:sheafprune_mechanism} in the main text shows two HR-Bench 8K examples from the existing visualization record, using $t=5$ and $\delta=1$. These settings differ from the default $t=3$ component table. The figure visualizes a local association between degree and corrected score; it does not establish question dependence or a universal failure mechanism.

\end{document}